\documentclass[10pt,final]{amsart}
\usepackage[utf8]{inputenc}
\usepackage[english]{babel}
\usepackage{graphicx}
\usepackage{amsfonts}
\usepackage{cite}
\usepackage{hyperref}
\usepackage[dvipsnames]{xcolor}

\begin{document}
\title[Efficient RNN methods for anomalously diffusing  trajectories]{Efficient recurrent neural network methods for anomalously diffusing single particle short and noisy trajectories}
\author{Òscar Garibo i Orts, Miguel A. Garcia-March, J. Alberto Conejero}
\address{Instituto Universitario de Matem\'atica Pura y Aplicada, Universitat Polit\`ecnica de Val\`encia, 46022 Val\`encia, Spain}
\email{osgaor@upv.es,garciamarch@mat.upv.es,aconejero@upv.es}%\author{S Else}
%address{ Other}
\date{\today}
\maketitle

\begin{abstract}
Anomalous diffusion occurs at very different scales in nature, from atomic systems to motions in cell organelles, biological tissues or ecology, and also in artificial materials, such as cement.  Being able to 
accurately measure the anomalous exponent associated to a given particle
trajectory, thus determining whether the particle subdiffuses, 
superdiffuses or performs normal diffusion, is of key importance to 
understand the diffusion process. Also it is often important to 
trustingly identify the model behind the trajectory, as it this gives a large amount of information on the system dynamics. Both aspects are particularly difficult when the input data are short and noisy trajectories. It is even more difficult if one cannot guarantee that the trajectories output in experiments are homogeneous, hindering the statistical methods based on ensembles of trajectories.  We present 
a data-driven method able to infer the anomalous exponent and to
identify the type of anomalous diffusion process  behind single, noisy and short trajectories, with good accuracy. This model was used in our participation in the Anomalous 
Diffusion (AnDi) Challenge. A combination of 
convolutional and recurrent neural networks was used to achieve 
state-of-the-art results when compared to methods participating in the 
AnDi Challenge, ranking top 4 in both classification and diffusion 
exponent regression.
\end{abstract}
%\keywords{Anomalous diffusion, Alabano-Kosovares}
%\submitto{\jpa}

%\begin{document}
%%%%%%%%%%

\section{Introduction}

Randomly moving particles, in some cases, diffuse anomalously in their surrounding medium. The concept of anomalous diffusion is defined in opposition to normal diffusion: since the movement is random, that is stochastic, the probability $P(\bf{x},t)$ of finding a particle at time $t$ and position $\bf{x}\in\mathbb{R}^d$, $d=1,2,3$ determines the dynamics. For normally diffusing particles, its width $\langle \bf{x}^2 \rangle $, known as Mean Squared Displacement (MSD), grows linearly with time. This occurs e.g. in the traditional  Brownian motion and is described by a partial differential diffusion equation (e.g. see a beautiful modern discussion on Fick,  Einstein, and Smoluchowski Diffusion Equations in \cite{2004IslamPS, 2004IslamPSb}). If the MSD does not grow linearly with time, that is  $\langle \bf{x}^2 \rangle\propto t^\alpha$, with $\alpha\ne1$, then the particles following such movement are said to anomalously diffuse in their medium.  The coefficient $\alpha$ is known as the anomalous diffusion coefficient. 

A great variety of systems can show anomalous diffusive behavior. Furthermore, the theoretical models best explaining such systems are also extremely heterogeneous. For example, some of the models describe particle motion as a sequence of displacements of random lengths occurring at stochastic times,  as in Brownian motion. Hence, both positions and times are stochastic variables whose behavior is determined by their corresponding Probability Distribution Functions (PDFs).  This behavior occurs in a wide class of models termed as continuous-time random walks (CTRWs)~\cite{1975ScherPRB}. A kind of  CTRW showing anomalous diffusion is that in which the PDF describing the stochastic times is a power-law distribution $\psi(t)\sim t^{-\sigma}$,  and displacements are sampled from a Gaussian PDF with variance $D$ and zero mean. Another class of models is obtained when, on top of a power-law PDF for the waiting times,  the PDF for displacements is not Gaussian \cite{1994klafterPRE}. These models are known as L\'evy walks, and here the stochastic times are known as flight times. For example, in one dimension the displacements  length is $|\Delta x|=|x_{i+1}-x_{i}|$, where $x_i$ is the position at time $t_i$, are correlated with the flight  times  the conditional probability  $\Psi(\Delta x | t) = \frac{1}{2}\delta(|\Delta x|-vt)$ where $v$ is the velocity. A model which results from the motion of a Brownian particle whose diffusion coefficient varies in time is the annealed transient time motion  (ATTM) model~\cite{massignan2014nonergodic}. Other models are obtaining considering a variety of situations and geometries, like the bouncing of a particle in a set of regions with partially transmitting boundaries of stochastic heights \cite{2019MunozFiP}, interactions between heterogeneous partners \cite{2017CharalambousPRE}, the movement of a particle in an environment with critical behavior \cite{2017MunozPRE}, etc. Another class of models can be defined from the Langevin equation: the stochastic differential equation governing the movement of a single particle with stochastic noise driving its movement (and modeling an environment interacting with the particle). Here, one may consider that the noise is non-white (termed as fractional Gaussian noise), with a normal distribution with zero mean but power-law correlations between the noise at different times. The resulting models are known as fractional Brownian motion  (FBM) models ~\cite{1968MandelbrotSR,Jeon2010Fractional}. Yet another class of models is obtained when, in the Langevin equation, one considers time-dependent diffusivity, even with white Gaussian noise~\cite{lim2002self}. This is known as scaled Brownian motion (SBM). For the anomalous diffusing case,  the diffusivity has power-law dependence with respect to $t$. See a review of anomalous diffusion models  in e.g.~\cite{2014MetzlerPCCP}. 

The anomalous diffusing behavior is diverse and, indeed, it can be best explained with many different theoretical models. The behavior is very different attending at the anomalous diffusion coefficient, $\alpha$. A limiting behavior occurs when  $\alpha$ is close to 0, as then the width of the PDF describing the position and times of particles does not change in time, being regarded as a trapping situation. If $\alpha$ lies in the interval $0<\alpha<1$,  the diffusion is called subdiffusive, while if $\alpha>1$, it is called superdiffusive. The limiting case of $\alpha=2$ is called ballistic motion and, of course, $\alpha=1$ corresponds to normal diffusion. 

Then,  a diffusing process of which one has access to the series of positions and times of a randomly moving particle, 
can be characterized by the anomalous diffusion coefficient and the model which betters explains its behavior. The tools which permit to do this characterization depend strongly on the availability of data. First, a possible situation is that we can guarantee the following two conditions in the experiment: i) a large quantity of long-enough trajectories can be obtained; ii) one can assure a homogeneity condition. This last condition means that all particles correspond to the same process over the whole experiment and can be assigned the same model and anomalous diffusion coefficient. In such a case, one can characterize the system performing an ensemble average between all trajectories~\cite{2015KeptenPlos1,2014MetzlerPCCP}.  A second possible situation is that one can assure the following conditions: i) one can obtain a very long trajectory; ii) one can assure that the particle's behavior does not change during the whole experiment; iii) one can assure that the behavior is ergodic, that is, that, with sufficient time, one realization of the experiment explores all possible configurations of the system. In such a case, one can use time averages to extract information of the process. But a third possible scenario is that in which the experiment is such that it may happen one or more of the following: i) the trajectories one can access are short; ii) one cannot assure that all trajectories of different particles are homogeneous; iii) one can only access to one or a few trajectories.  In such case, to assign a single short trajectory to a diffusion process characterized by a theoretical model and an $\alpha$ one has to find alternative tools to ensemble and time averages of high-quality data. One possible route is to use an approach based on a machine learning tool. In this paper, we present a high accuracy tool based on a particular kind of artificial recurrent neural network that has shown its utility and good performance when dealing with time series: the Long short-term memory (LSTM) architecture~\cite{1997HochreiterNC,2015Liptonarxiv}.

Large theoretical efforts have built a battery of statistical techniques to find out the anomalous exponent given these difficulties. A non-comprehensive list includes Ref.~\cite{2010TejedorBJ} where a method based on the mean maximal excursion method was proposed, Refs. ~\cite{2011MakaravaPRE,2016HinsenJCP} where a Bayesian estimation was proposed and tested for FBM processes, Ref.~\cite{2015BurneckiSciRep} where a method based in a fractionally integrated moving average was introduced, and Ref.~\cite{2018KrapfNJP,2019KrapfPRX} where a method based in the information contained in the power spectral density of a single trajectory was proposed (for reviews and other methods see also~\cite{2013KeptenPRE,2019WeronPRE,2015MerozPR}).  Also, a Bayesian approach to test among different types of motion, which includes free motion (normal diffusion) and subdiffusion, was proposed in  \cite{2012MonnierBJ}.  This paper considers other two types of motion: (i) {\it confined diffusion}, where particles cannot exit some structure, say a sphere in three dimensions \cite{2007BickelPhysA}; and (ii)  {\it directed motion}, where there is some flow in the ensemble of Brownian particles, for example, due to Brownian motors \cite{2002AstumianPT}, which results in a ballistic MSD. On the other hand, statistical methods have been used to distinguish among models. For example in~\cite{2009MagdziarzPRL,2010JeonJPA} methods were introduced to distinguish among FBM and CTRW; in~\cite{2012BurneckiBJ} an algorithm to identify and characterize FBM was introduced; Ref.~\cite{2018ThapaPCCP} presents a Bayesian method to distinguish among Brownian motion, SBM and FBM (see also \cite{2019CherstvySM}); finally in Refs.~\cite{2013MerozPRL,2017ChenPRE} it is discussed a method to distinguish among different physical origins for subdiffusion, which in turn point out to the different possible theoretical models.

Very recently, there has been a sudden growth of proposals that face this same problem with machine learning tools.  A random forest classification algorithm was used to distinguish among directed motion, normal and anomalous diffusion was introduced in \cite{2016WagnerBP} and extended to include confined motion in \cite{2017WagnerPlos1}. A random forest was also used to classify trajectories as CTRW, ATTM, FBM, and LWs in \cite{2020MunozNJP} and also to assign an $\alpha$ single trajectories (see also \cite{2020JanczuraPRE,2020LochEntropy} where random forest and gradient boosting is used to classify among normal, super- and subdiffusion).  In \cite{2019BoPRE} a recurrent neural network was used to extract the exponent from a single short trajectory, even when the trajectory is sampled at irregular times. Also, in \cite{2021ArgunArxiv} a recurrent neural network is used to classify between the five models described above (CTRW, FBM, ATTM, LW, SBM) and obtain the anomalous exponent.  In \cite{2020HanELife} a recurrent neural network was used to estimate the Hurst exponent of an FBM. A set of convolutional neural networks used to classify among Brownian motion, FBM, and CTRW, with simultaneous estimation of  Hurst exponent $H$ (which is related to the anomalous exponent $\alpha$ as $H=\alpha/2$) for FBM and the diffusion coefficient for Brownian motion, was presented in \cite{2019GranikBJ}. A convolution neural network was also used in  \cite{2019KowalekPRE} to classify trajectories as normal diffusion, anomalous diffusion, directed motion, or confined motion, and compared with random forest and gradient boosting. A combination of classical statistics analysis with  supervised deep learning (a deep feed-forward neural network to cluster parameters extracted from the statistical features of individual trajectories) was used to classify among  FBM, ATTM, CTRW, SBM, and LWs,  and infer $\alpha$ was introduced in \cite{2021GentiliArxiv}. 
A recent review on machine learning in the nearby field of active matter can be consulted in \cite{2020CichosNMI}

This research effort was the reason why the Anomalous Diffusion (AnDi) Challenge was launch in March 2020 (http://www.andi-challenge.org) \cite{2020MunozArxiv,2021munoz-gilArx}. Similarly, as References  \cite{2021ArgunArxiv,2021GentiliArxiv}, the research described in this paper was a response to this challenge. In this challenge of the tasks were, for short, noisy trajectories either in one-, two- or three dimensions: i) to propose and test a method able to distinguish among FBM, CTRW, ATTM, SBM, and LWs; ii) to propose and test a method to get the anomalous exponent. Here we present the tool which performed among the best ones in all these tasks and was best in the first task in one dimension. After a brief discussion of the experimental context for anomalous diffusion in next subsection, we describe in Sec. \ref{sec:method} the details of the method. In Sec. \ref{sec:results} we present the results obtained for both tasks, that is, inference of anomalous diffusion and classification according to theoretical model. We offer our conclusions in Sec. \ref{sec:conc}.

\begin{figure}[h]
\centering
\includegraphics[width=0.8\columnwidth]{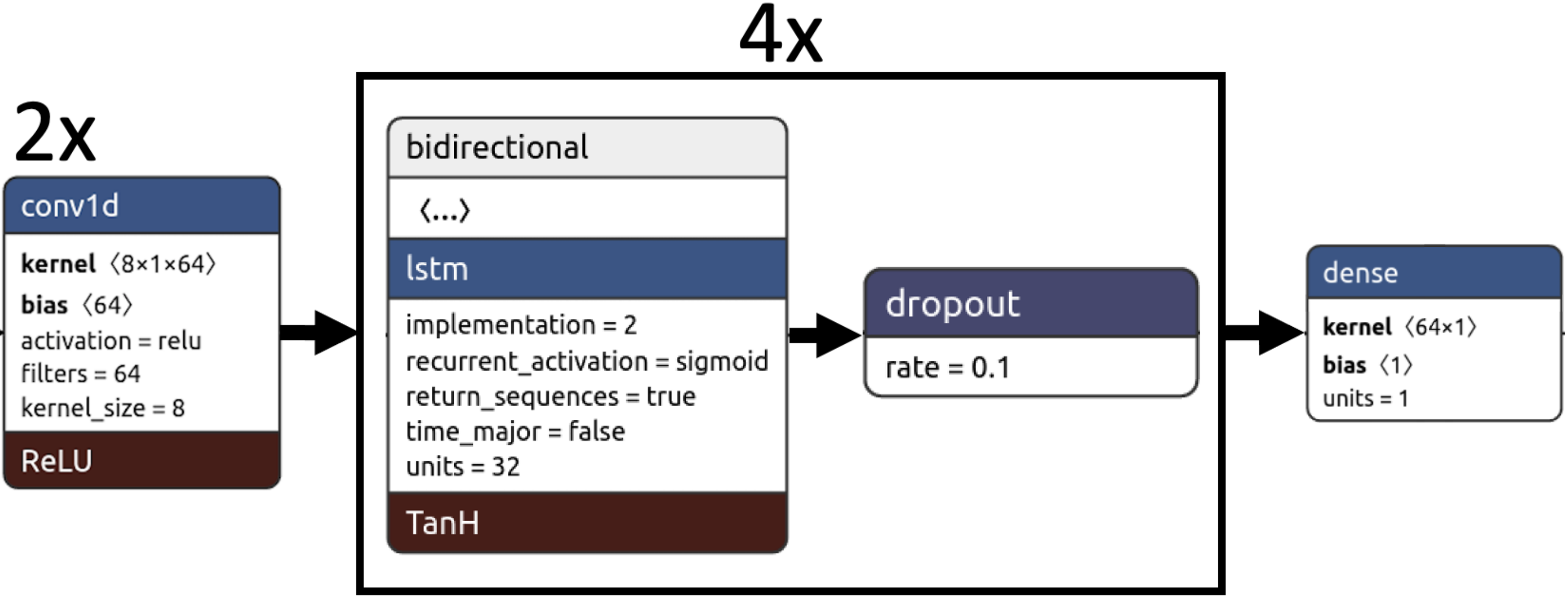}\\
\vspace{0.5cm}
\includegraphics[width=\columnwidth]{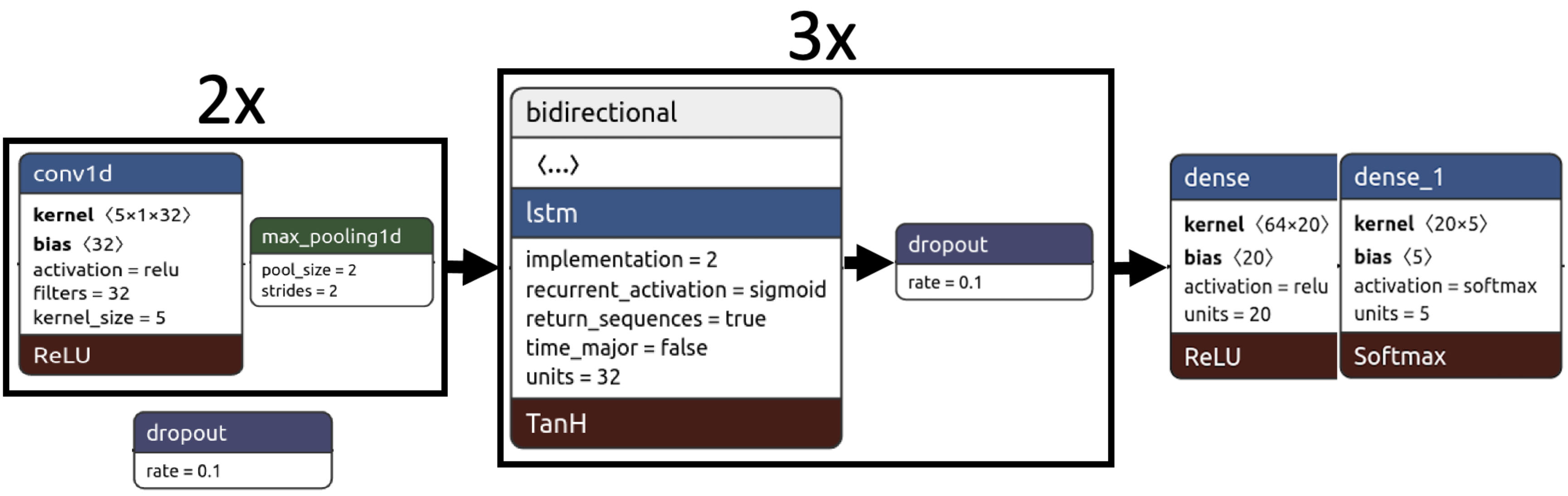}
\caption{{\it Schematic of the methods}. Top: Schematic of the method used for regression.
Down: Schematic of the method used for classification.}
\label{fig:fig1}
\end{figure}

\subsection{Anomalous diffusion in experiments}

Anomalous diffusion occurs in a plethora of experimental situations, ranging all scales \cite{2019OliveiraFiP,2005KlafterPW}. The smallest scale we are aware of occurs at the level of atoms, particularly in experiments with ultracold atoms \cite{2012SagiPRL,2019DechantPRL,2017kindermannNatPhys} and also of quantized vortices in Bose-Einstein condensates \cite{2021TangPNAS}. Also, examples exist for ions in solutions\cite{2013LenziIJES}. Many experiments show anomalous diffusion in biological systems. For example, transient anomalous diffusion occurs for telomeres motion in the nucleus of cells~\cite{2009BronsteinPRL,2017stadlerNJP,2019KrapfPRX}. Generally speaking, the recent developments in single-particle tracking techniques have boosted a revolution in cell biology~\cite{2015ManzoROPP}, and several experiments have found anomalous diffusion, e.g. in the plasma membrane~\cite{2011WeigelPNAS,2015ManzoPRX} or in the cytoplasm~\cite{2000CaspiPRL,2010WeberPRL,2013RegnerBJ,2020SabriPRL}. Also, anomalous diffusion occurs in larger systems, like in worm-like micellar solutions~\cite{2013JeonNJP}, yeast cells~\cite{2004TolicPRL},  water in  porous biological tissues~\cite{1996KopfBJ,2006OzarslanJMR,2013MaginMMM}, in cement-based materials~\cite{2020ZhangTPM}, or ecology (see e.g. ~\cite{2016Mendez}). 

The characterization of the kind of model that better explains the data obtained in an experiment and the associated anomalous exponent takes on key importance in many of these systems. For example, there has been large discussion on the underlying diffusion model and ergodicity  which occurs in the experiments in \cite{2006GoldingPRL,2004TolicPRL} (see~\cite{2008HePRL,2009MagdziarzPRL,2011MagdziarzPRE,2016MolinaPRE}). Also, since diffusion is the central transport mechanism in biological cells, if it is anomalous, it impacts how the system works. For example, it has been discussed that it may have an impact in chemical reactions \cite{2018LanoiseleeNatCom}.   Also, anomalous diffusion is compatible both with ergodic behaviors and non-ergodic behaviors, where a single realization does not explore all possible configurations (realizations) of the system. In the context of diffusion, one has weak ergodicity breaking if the averages taken over a single realization in infinite time do not equal ensemble averages over many realizations~\cite{2014MetzlerPCCP}. Also, for ultra-weak ergodicity breaking time and ensemble averages differ by a constant factor~\cite{2013Godec,2013Godecb}.  Processes like CTRW, ATTM and SBM show weak ergodicity breaking \cite{2015ManzoPRX,massignan2014nonergodic,2005BelPRL}, whereas Brownian motion and FBM are ergodic \cite{2009DengPRE,2017SchwarzlSciRep}.  This is thus an important feature that may mark the ability to distinguish between models. Experimental signals are always noisy (e.g. in single-particle tracking, due to localization precision~\cite{2014ChenouardNM}). Often noise hides non-ergodic behavior~\cite{2013JeonJCP} and hinders statistical analysis.  Also, experimental trajectories are often short, depending of the kind of experiment \cite{2020MunozBA}. Finally, one may not be able to assure that, in a biological system, that the measure trajectories are homogeneous. Therefore, a tool able to characterize diffusion from a single trajectory which is the output of an experiment, which is short and noisy, will find a great utility in a plethora of applications. 

\section{Description of the method}\label{sec:method}
\subsection{Generation of training and validation sets}

The performance of a supervised machine-learning technique depends greatly on the quality of the data used to train. In this subsection, we discuss how we designed the training and validation sets, with the goal to have an homogeneous enough training set, that is, containing enough instances of all five models and a variety of anomalous coefficients, $\alpha$, and trajectories lengths,  yet being not too big, and hence computationally tractable.

To build the training and validation data set, we used the code provided in the AnDi Challenge \cite{2020MunozArxiv,2021munoz-gilArx} that is available at \url{https://github.com/
AnDiChallenge} on GitHub.  We thus considered trajectories generated by the five classes (ATTM, CTRW, FBM, LW and SBM), with lengths ranging from 10 to 1,000. Considering steps of size 10, it makes 100 possible different trajectory lengths. Taking into account that in some models, a change in the regime is likely to show at some point in the trajectory, multiplying by a factor of 100 we can ensure at least one trajectory of each type in each part of the trajectory.
Not all models can cover the whole range of diffusion coefficients. We considered values of $\alpha$ in the range of 0.02 and 1.95 with increments of 0.05, which means 19 different possible values for $\alpha$. Note that the resulting set is imbalanced with respect to the models (e.g. CTRW is more represented than FBM in the subdiffusive regime, and LW is more represented than FBM in the superdiffusive regime).
Putting all together, a data set of size $10^6$ to ensure that, on average, at least one trajectory of each possible type will be present in the data set. We also consider trajectories with signal-to-noise ration (SNR) equal to 1 and 2, so this increases the recommended dataset size up to $2\cdot 10^6$. We use then sizes equal or larger than this number. As it can be seen in \cite[Fig.2]{2021munoz-gilArx} the performance of the models is not improved if we increase the SNR from 2 to 10.

In Table \ref{tab:tab1} we show the number of trajectories used for regression. In both tasks, we split available training data into training (90\%) and validation (10\%) independently at each epoch. The models are trained until no improvement was achieved after 10 consecutive epochs. We point out that when training and testing models were imbalanced. The reason is that when testing our models against the validation set provided by the organizers and some other validation datasets generated by us, we notice an improvement in the performance with the size of the training dataset. Moreover, it is worth mentioning that the validation dataset provided by the organizers contained only 10k trajectories, and in our case, our validation datasets were at least of size 200k trajectories.

\begin{center}
\begin{table}
\begin{tabular}{|l|l|c|c|c|}
\hline
 Task & Trajectory length & 1D   &  2D   & 3D    \\
\hline
Regression & $L\in [10, 20[$ & 8 & 4  & 4   \\
Regression &  $L\in [20, 50[$ & 18 &  12 & 12   \\
Regression &  $L\in [20, 100[$& 4 & 3 &  3 \\
Regression &  $L\in [100, 1000]$& 2  &2  & 2 \\  
Classification & $L\in [10, 1000]$ & 4 & 2 & 2\\
 \hline
\end{tabular}
\caption{\label{tab:tab1} {\it Number of trajectories per task, dimension, and length used as data set.} Number of trajectories scaled in millions, $\times10^6$.}
\end{table}
\end{center}

\subsection{Architecture of the method}

The basic architecture used both for classification and regression consists of two convolutional layers used to extract spatial features from the trajectories. An initial convolutional layer is set with 32 filters with a kernel size of 5, making a sliding window of size 5 which slides through each trajectory extracting spatial features from them.

A second convolutional layer is used with the number of filters increased to 64 to extract higher-level features. Depending on the task, we reduce the dimensionality by applying a maxpool layer (in the classification task, not in the regression task) after each convolutional layer. The resulting encoded trajectories are fed in three stacked bidirectional LSTMs to learn the sequential information, with a drop-out layer of the 10\% of the nodes to avoid incurring into overfitting. Finally, we use several fully connected dense layers to predict the desired information (exponent regression or model classification).

\subsubsection{Particularities of the method used for regression } 

We have used a trajectory length dependent approach by building models for different trajectory lengths. The following bins have been used based on trajectory length: $[10,20]$, $]20,30]$, $]30,40]$, $]40,50]$, $]50,100]$, $]100,200]$, $]200,300]$, $]300,400]$, $]400,500]$, $]500,600]$, $]600,800]$ and $]800,1000]$, what makes a total of 12 different models, all sharing the same architecture. 
We have two convolutional layers followed by 4 bidirectional LSTM blocks. After each block, a dropout layer is set. The output of the last one feeds a one node fully connected dense layer with linear activation function to get the estimated diffusion exponent (see Fig. \ref{fig:fig1}, top panel).

\subsubsection{Particularities of the method used for classification }

Here we use a single model for all possible trajectories lengths and apply lead padding to each trajectory to make them of the same length (1,000). During the experimentation we found that applying dimensionality reduction layers (maxpool) at the output of each convolutional layer helped the LSTM layers to extract better sequential information to classify the trajectories, since doing so allowed to reduce the level of noise in the extracted features. The output of the last LSTM layer feeds a fully connected twenty-nodes dense layer with Relu activation function to capture non-linearity. This layer is followed by the final five nodes dense layer with softmax activation function to obtain five different probabilities for each trajectory to belong to one of the five possible models (see Fig. \ref{fig:fig1}, right panel).

\section{Results}\label{sec:results}
\subsection{Inference of the anomalous diffusion exponent}

Here we present the results for one dimensional trajectories, while this tool was also used in two and three dimensions. The results in two and three dimensions are qualitatively similar, so for clarity and brevity we choose to discuss only the one dimensional-case. The tool is available in the web site of the AnDi Challenge \url{http://andi-challenge.org/} and therefore results in all dimensions can be accessed there. Additionally, the code is available in \url{https://github.com/AnDiChallenge/AnDi2020_TeamM_UPV-MAT}.

 To evaluate the accuracy of our results, we calculate the Mean Absolute Error (MAE) between predicted numerical $\alpha_{\rm{num}}$ and the ground truth value $\alpha_{\rm{GT}}$. For $N$ trajectories in the test set we compute the MAE as
 
\begin{equation}
\rm{MAE}=\frac{1}{N}\sum_{j=1}^N|\alpha_{j,\rm{num}}-\alpha_{j,\rm{GT}}|,
\end{equation}
where the subindex $j$ refers to the $j$-th trajectory. In the test set we included $N=2000$ trajectories of increasing length   $L\in \{20,30,40,50,100,200,300,400$, $500,600,800,1000\}$, resembling the binning carried out in the training (see Figures). 

\begin{figure}[h!]
\centering
\includegraphics[width=0.49\columnwidth]{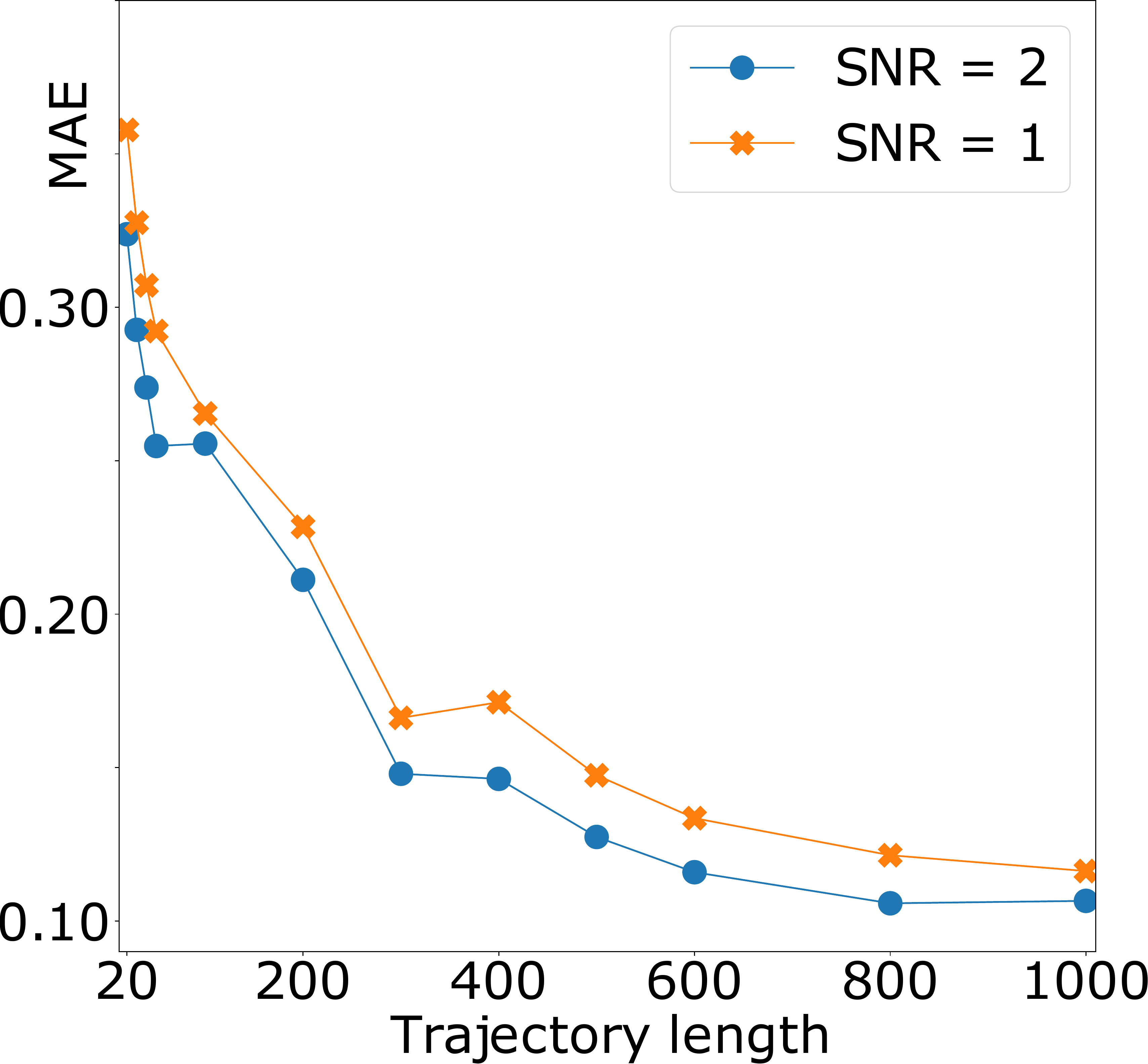}
 \caption{\label{fig:fig2} {\it Inference of the anomalous exponent $\alpha$ as a function of length.}  MAE as a function of length for  $\rm{SNR}=2$ and   $\rm{SNR}=1$  (that is $\sigma=0.5$ and $\sigma=1$).}
\end{figure}

We used a pool of models: trajectories generated with a CTRW, FBM, LW, SBM, and ATTM. The trajectories produced with ATTM and CTRW are subdiffusive, i.e.  $0<\alpha\le 1$ while the trajectories produced with LWs model are only superdiffusive, i.e. $1\le\alpha\le 2$. Those produced with SBM and FBM cover the whole range of anomalous exponents, $0<\alpha\le2$. Each trajectory is corrupted with some noise. To this end we consider the standard deviation of the displacements $\sigma_{\rm{D}}$ and add some Gaussian noise, with a standard deviation $\sigma_{\rm{noise}}$, which is some portion of the $\sigma_{\rm{D}}$. The SNR is thus $\rm{SNR}=\sigma_{\rm{D}}/\sigma_{\rm{noise}}$. We evaluate moderate and high noise, that is $\rm{SNR}=2$ and  $\rm{SNR}=1$, which therefore means that $\sigma_{\rm{noise}}$ is half of $\sigma_{\rm{D}}$ or coincides with it. In Fig.~\ref{fig:fig2} we plot the MAE as a function of length for different lengths  of the trajectory and the two different noise levels. 

\begin{figure}
%\begin{centering}
\centering
\includegraphics[width=0.49\columnwidth]{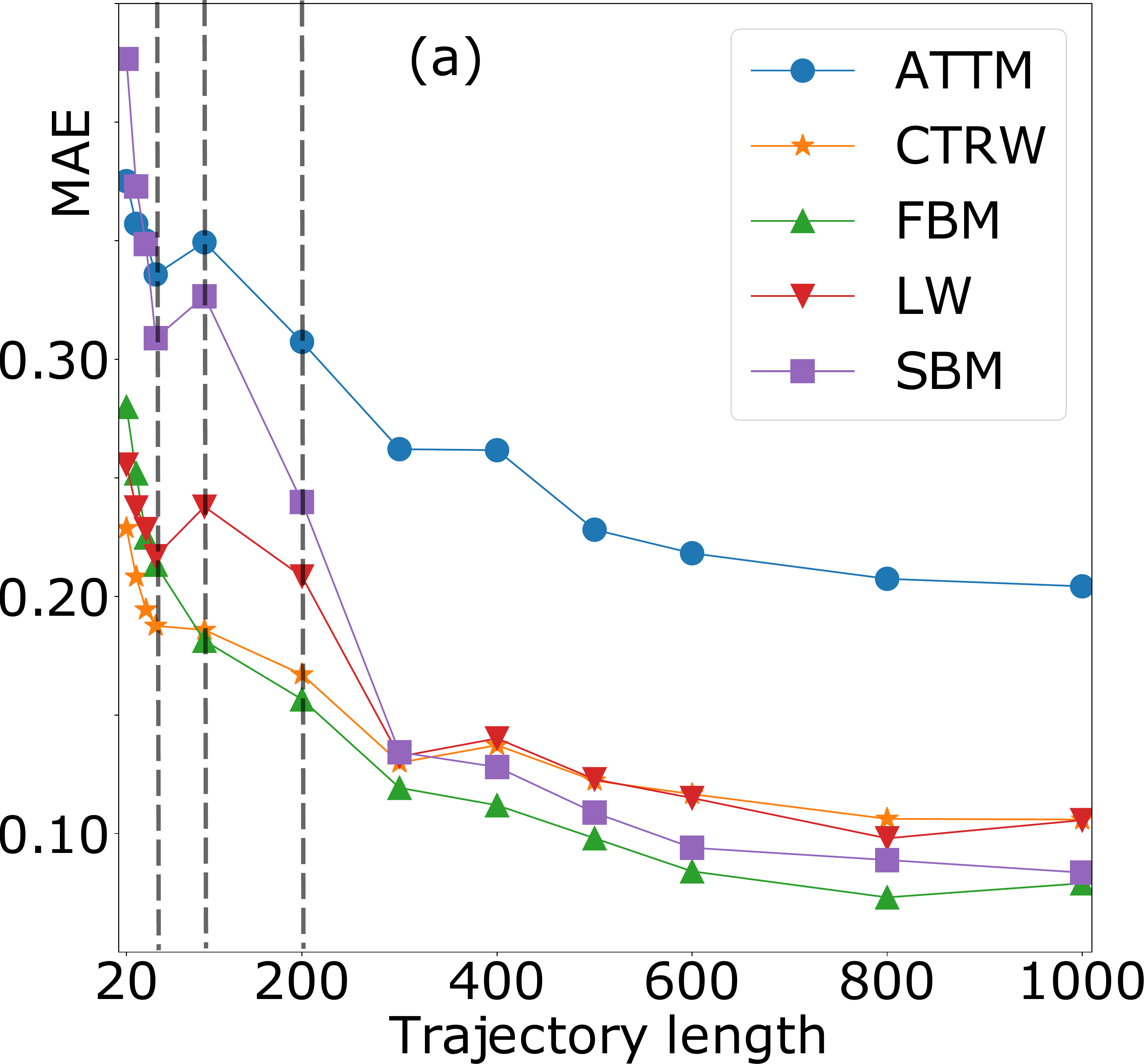}
\includegraphics[width=0.49\columnwidth]{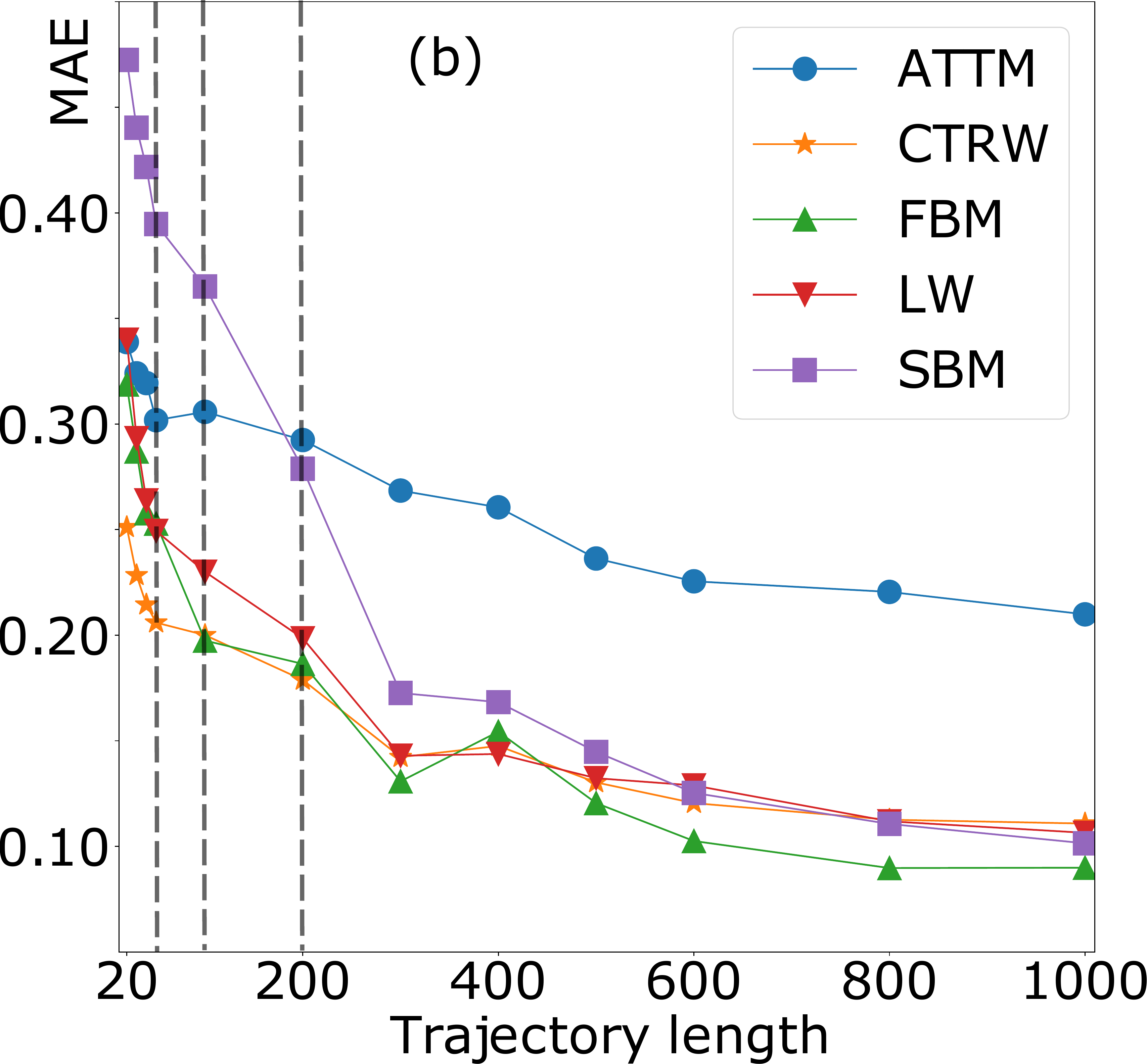} 
 %\end{centering}
 %
 % \centering
 \caption{\label{fig:fig3} {\it Inference of the anomalous exponent $\alpha$ as a function of length for different theoretical models.}  MAE as a function of length for (a) $\rm{SNR}=2$ and  (b) $\rm{SNR}=1$. Vertical dotted lines signal results at $L=50, 100, 200$, to help discussion. }
\end{figure}

As expected, the MAE gets better as the trajectories get larger. Mean absolute error improvement stabilizes around $L=500$. Also as expected the results for  $\rm{SNR}=2\,(\sigma=0.5)$  are better than for   $\rm{SNR}=1\,(\sigma=1)$, for all lengths. For the length $L=20$, with the current architecture, MAE between $0.3$ and $0.45$ are reached, which we set as a validity limit of the model. In Fig. \ref{fig:fig3} we plot the MAE as a function of length for the different models and those above two different levels of noise. 

The results show an abrupt change in accuracy  in the range $20<L<300$ for the SBM model for both levels of noise. For short trajectories ($L<50$) and lower level of noise, $SNR=2$, ATTM and SBM behave similarly but worse than CTRW, FBM, and LW. Surprisingly, for larger noise ($SNR=1$) and short trajectories ($L<50$), ATTM have similar performance as the rest of models, while SBM keeps performing worse than any model. Nevertheless, ATTM reduces MAE as $L$ is increased quite slowly ($L>50$). Conversely, FBM, LW and CTRW quickly improve their MAE for short trajectories ($L<50$), showing already reasonably good MAEs starting at lengths around  $L=100$.  Finally, since ATTM,  FBM, LW and CTRW show quite stable MAEs in the range   $100<L<300$, we note that  most of MAEs change in Fig.~\ref{fig:fig2} in this range is due to the trajectories generated with the SBM model. 

\begin{figure}
 \includegraphics[width=0.49\columnwidth]{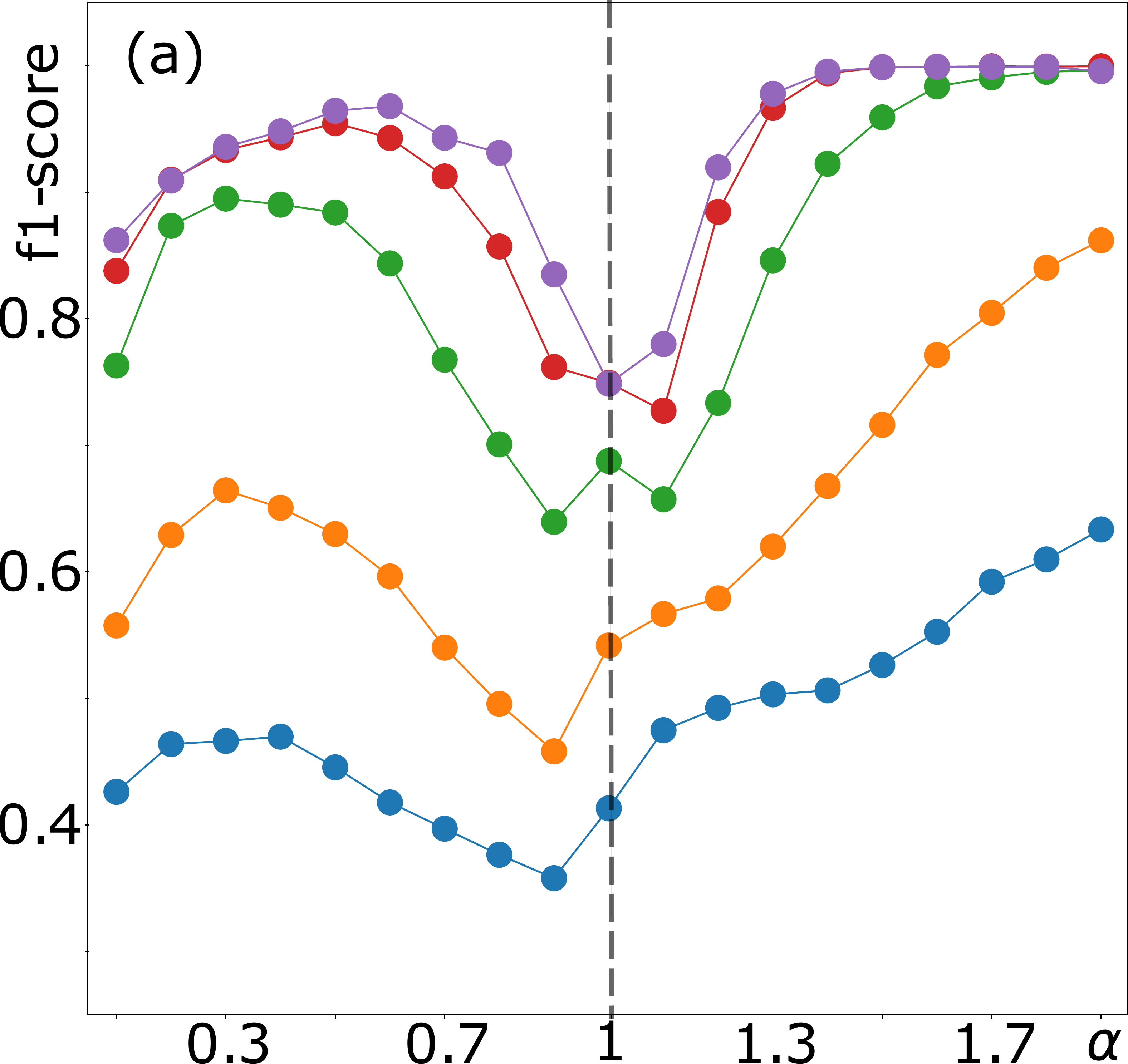} 
 \includegraphics[width=0.49\columnwidth]{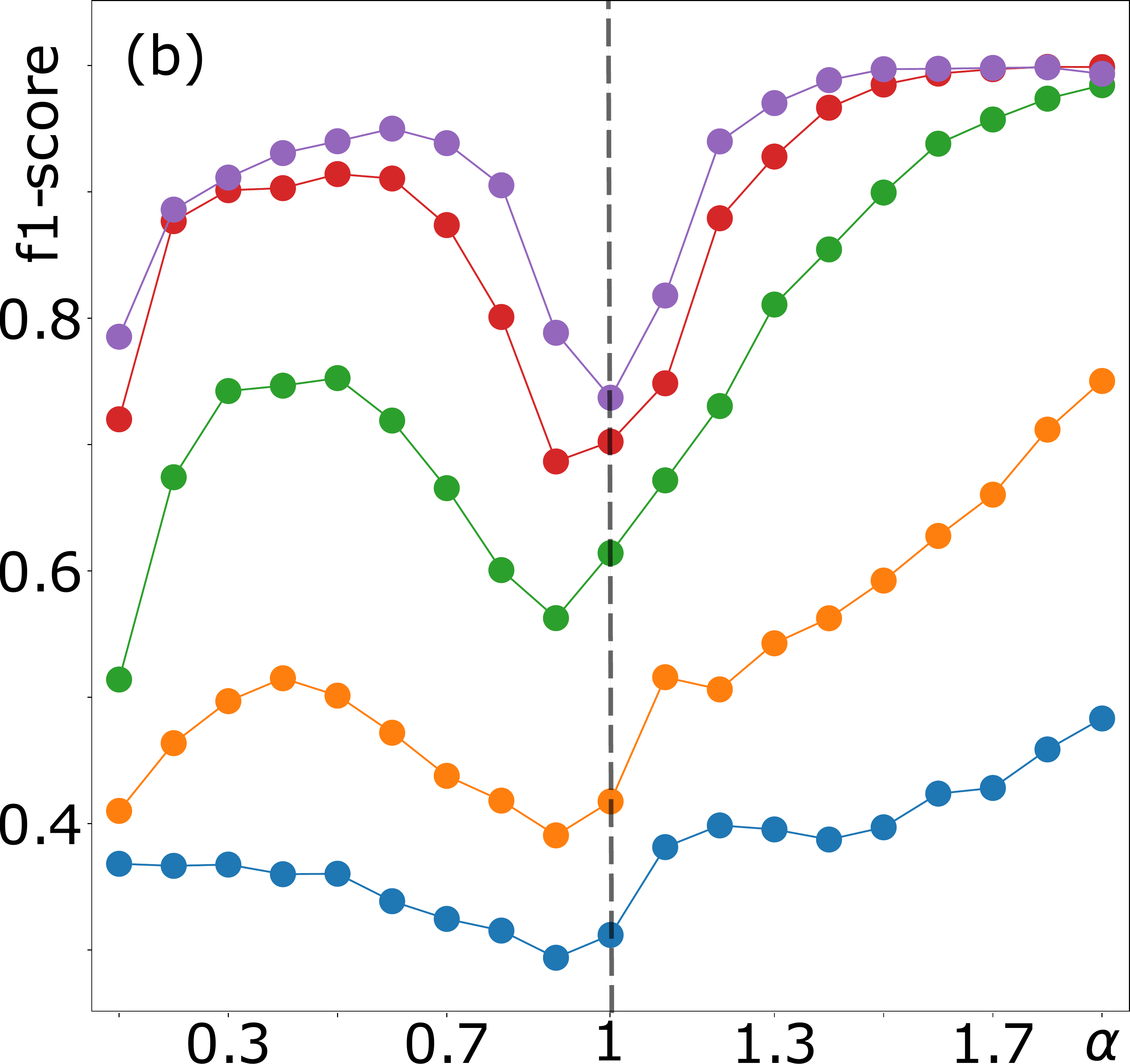}
 \includegraphics[width=0.49\columnwidth]{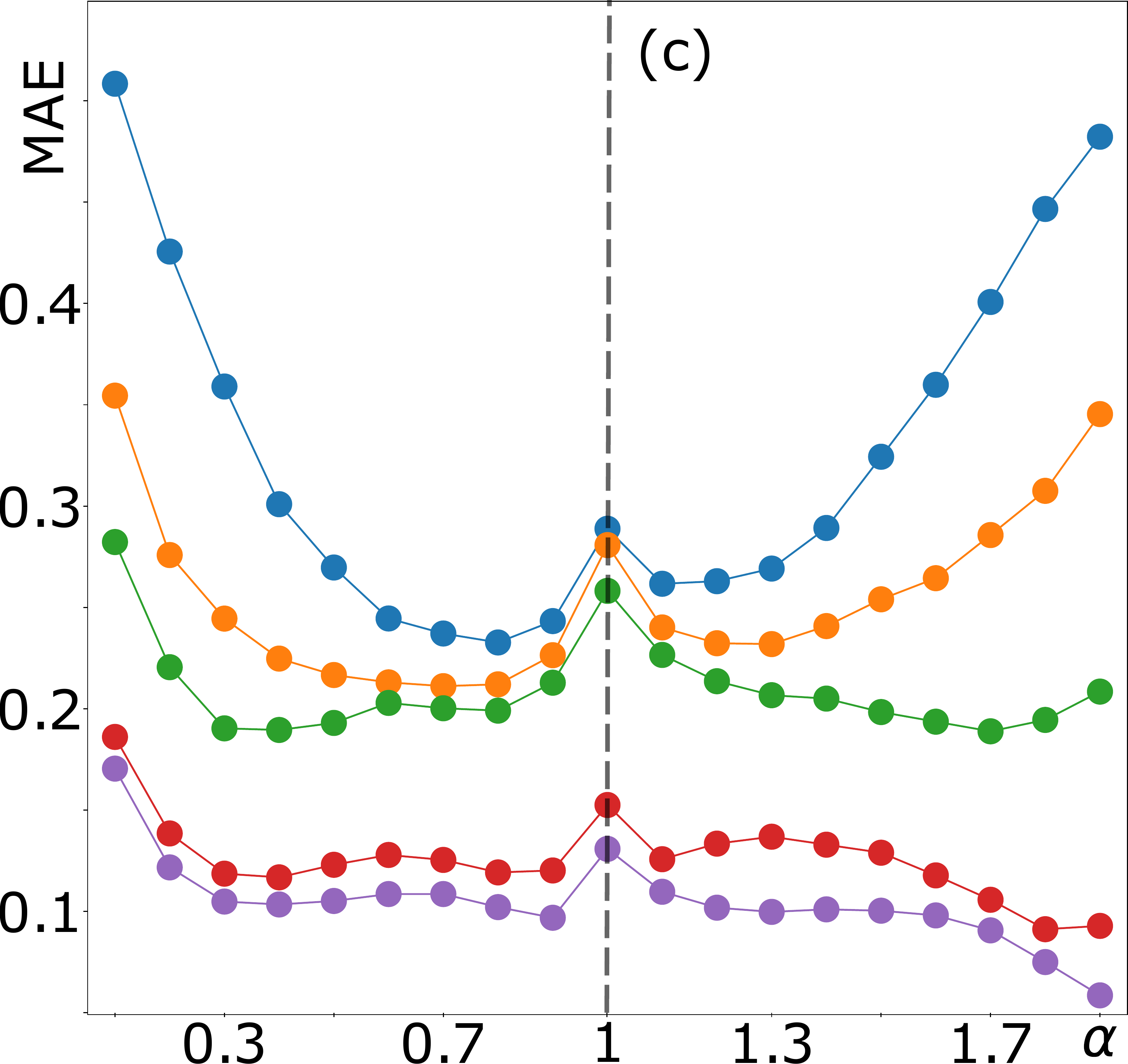} 
 \includegraphics[width=0.49\columnwidth]{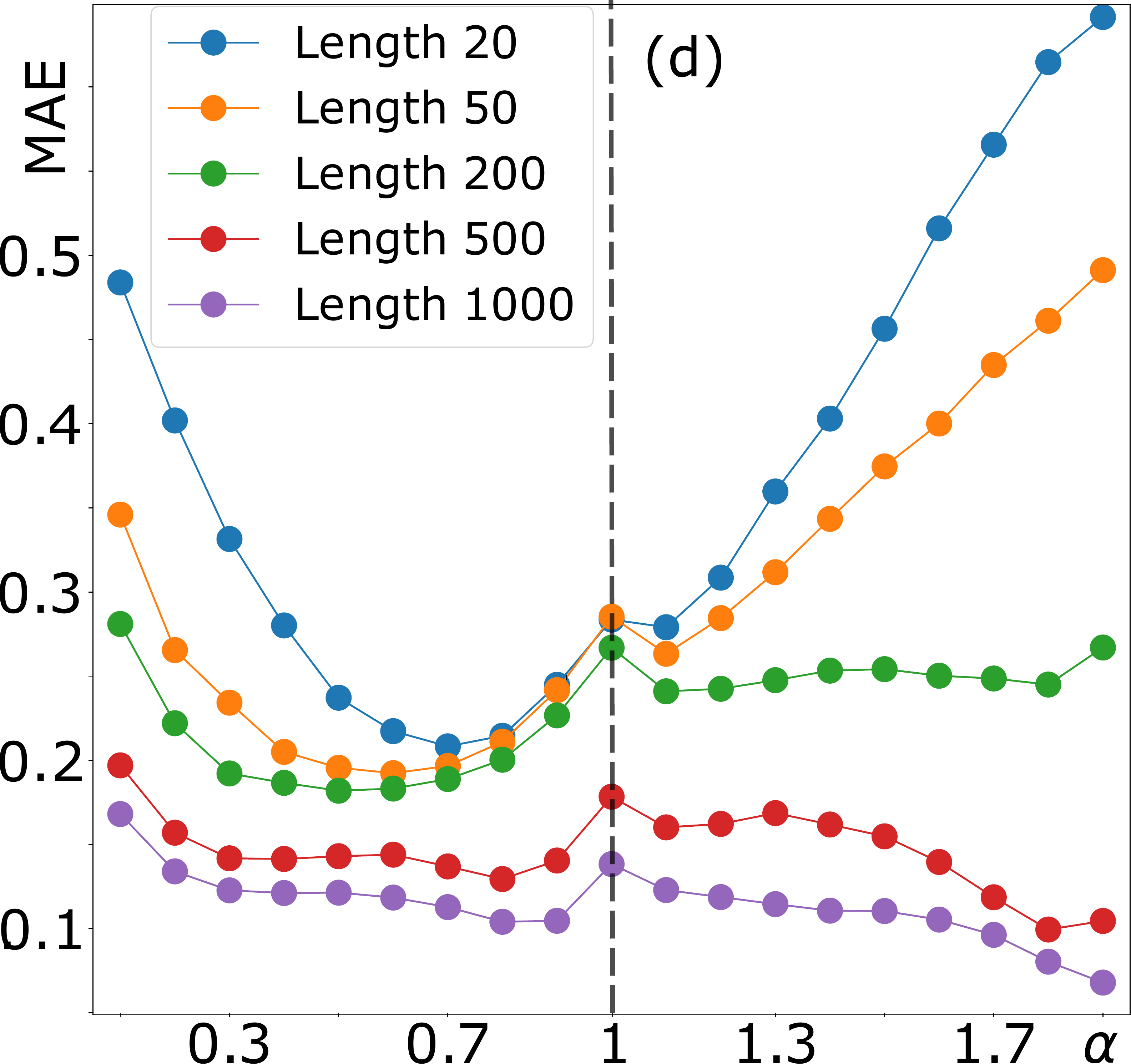}
% \centering
 \caption{\label{fig:fig4} {\it Inference of the anomalous exponent $\alpha$ for different values of the ground truth.}    f$_1$-score and MAE for  $\rm{SNR}=2$  (a) and (c) and $\rm{SNR}=1$  (b) and (d), for $L= 20, 50, 200, 500$ and $1000$ (see legend) and trajectories in one dimension. Dotted vertical line signals normal diffusion. }
\end{figure}

It is interesting to fix length, and have a closer look on how the model works for different values of the anomalous exponent. This is what we show in Fig.~\ref{fig:fig4}.  Here, given a length $L$ and a SNR value, we perform calculations for ground truth values of the anomalous exponent in the interval $\alpha_{\rm{GT}}\in[0.1,1.9]$, in discrete increments of $\Delta \alpha=0.1$. Here, we have also calculated the  f$_1$-score, which is defined as
$\rm{f}_1=\rm{TP}/N$,
that is, the ratio of true positives (TP) over the total number of trajectories in the test set, $N$. We consider a  TP when the predicted value of $\alpha$, $\alpha_{\rm{num}}$,  lies in the interval   $[\alpha_{\rm{GT}}-\Delta \alpha/2,\alpha_{\rm{GT}}+\Delta \alpha/2]$.  

\begin{figure}
 \includegraphics[width=0.5\columnwidth]{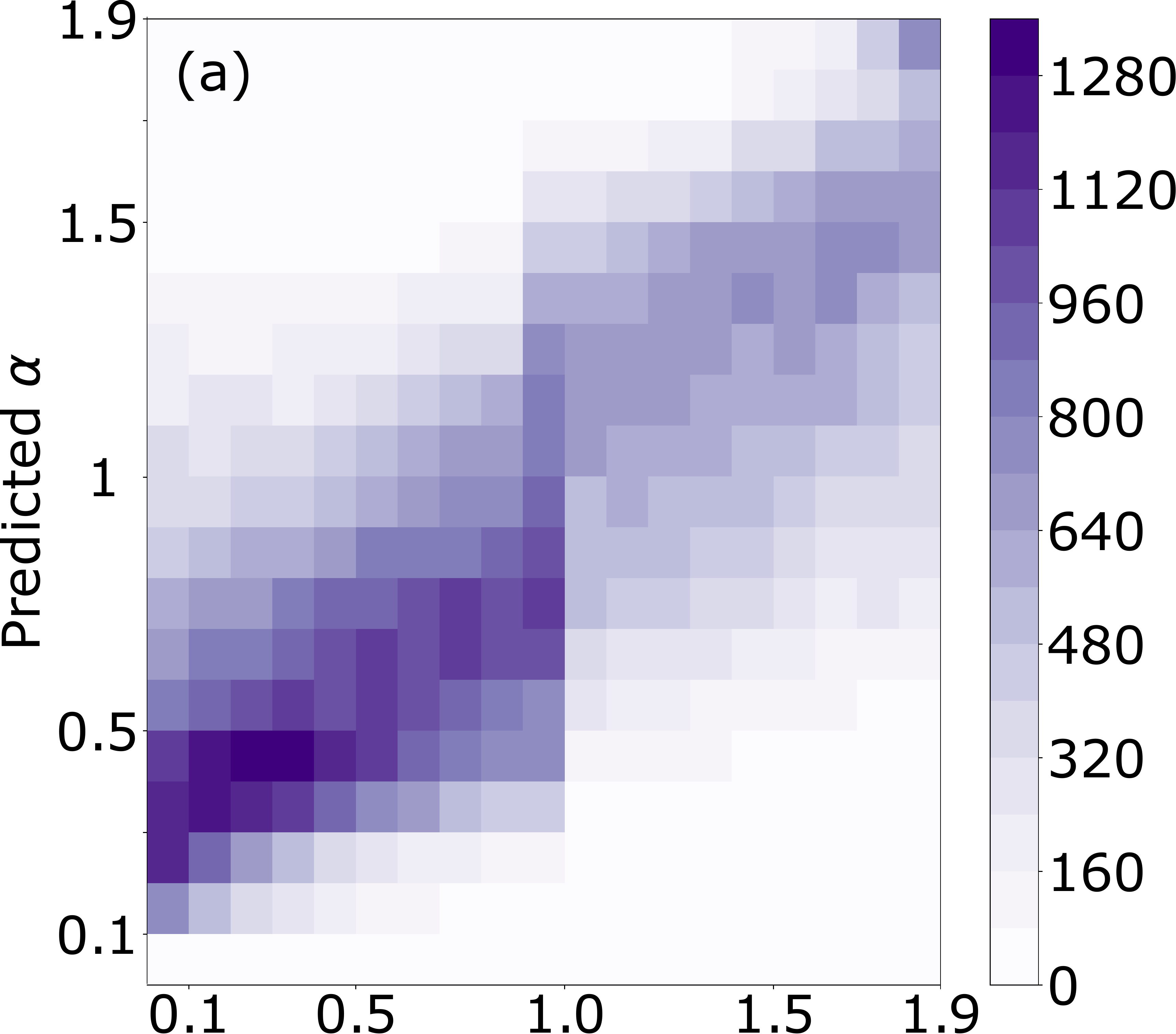}
 %\centering
 \includegraphics[width=0.48\columnwidth]{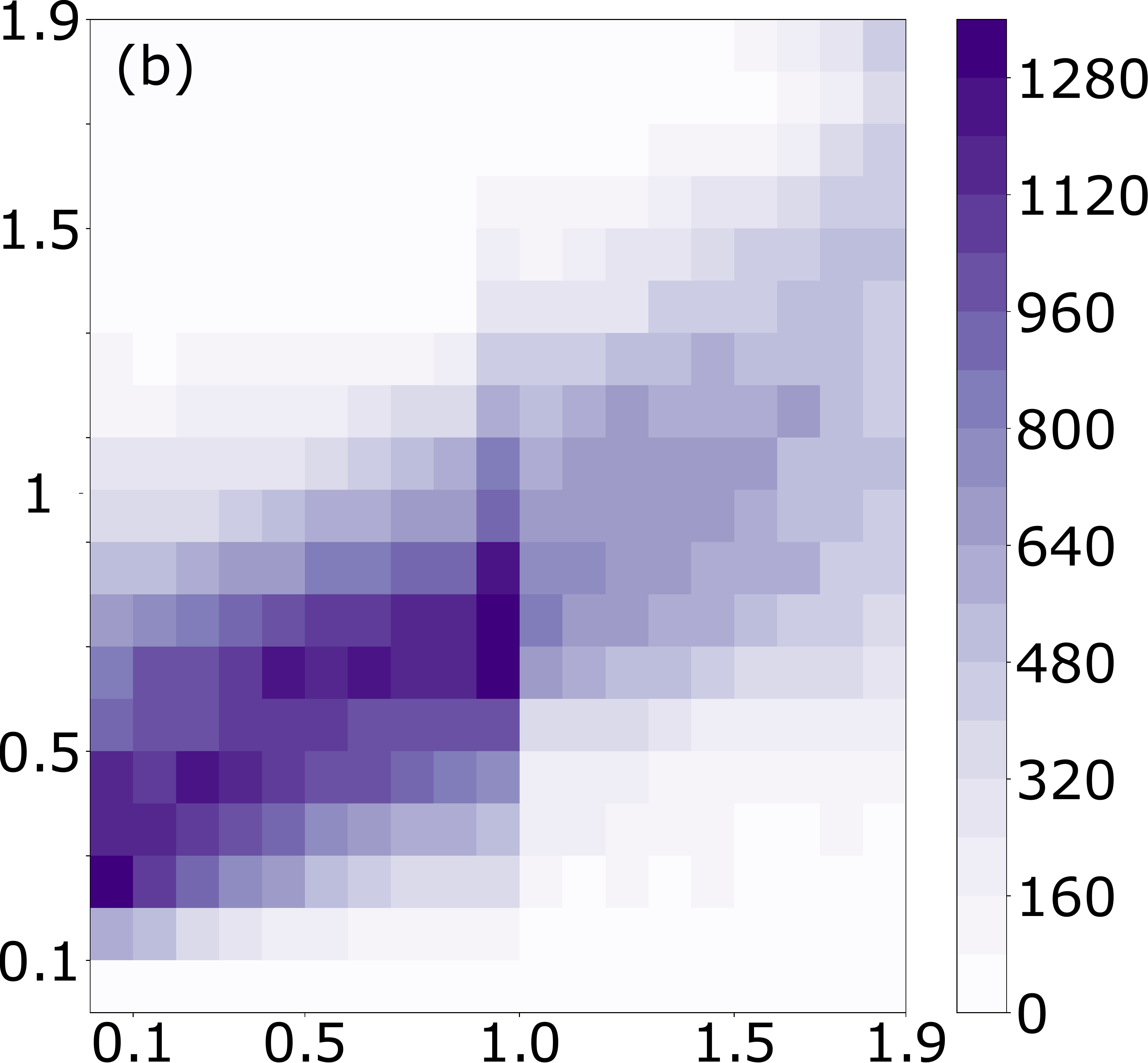}\\
 \includegraphics[width=0.5\columnwidth]{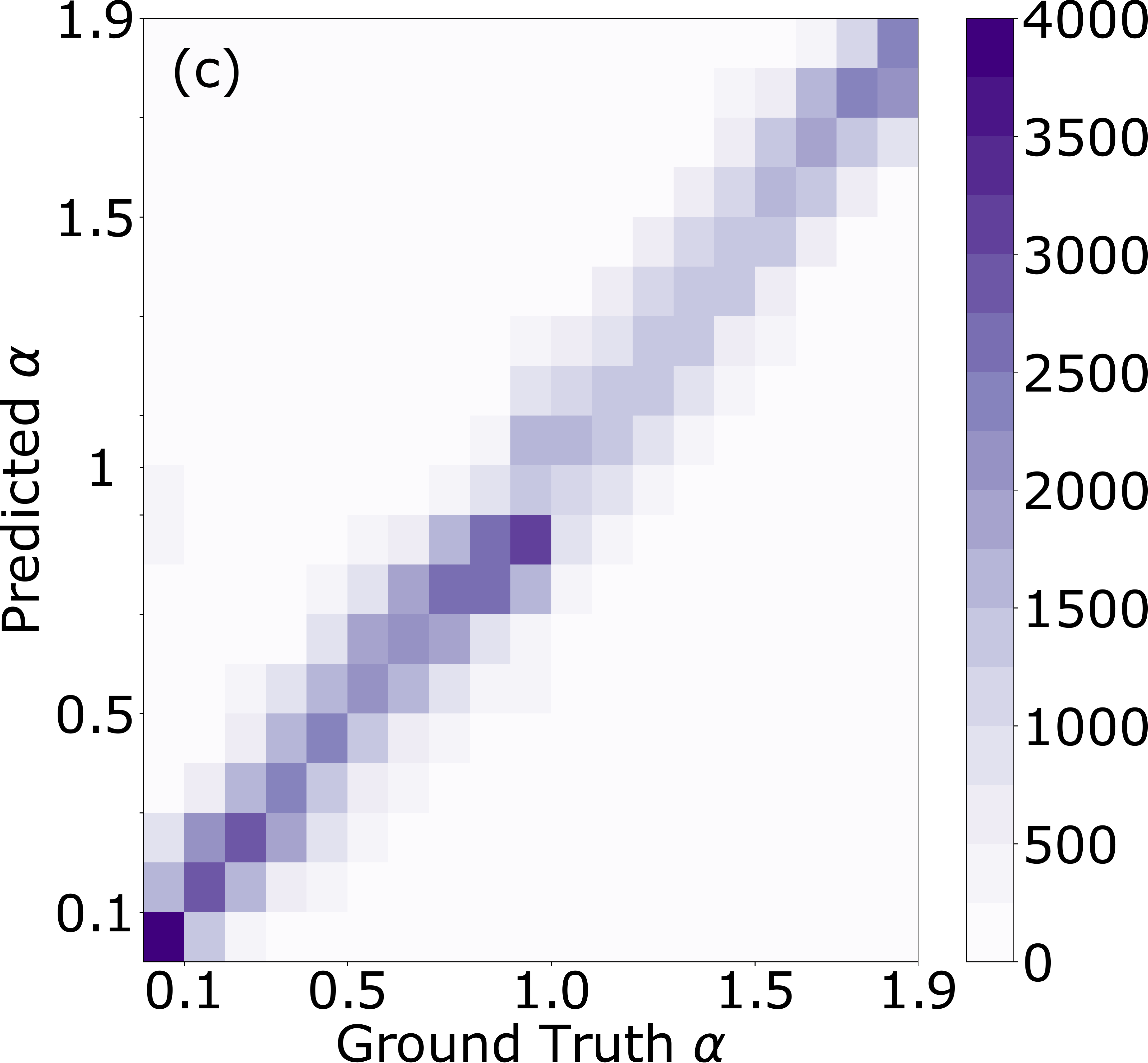}
 %\centering
 \includegraphics[width=0.48\columnwidth]{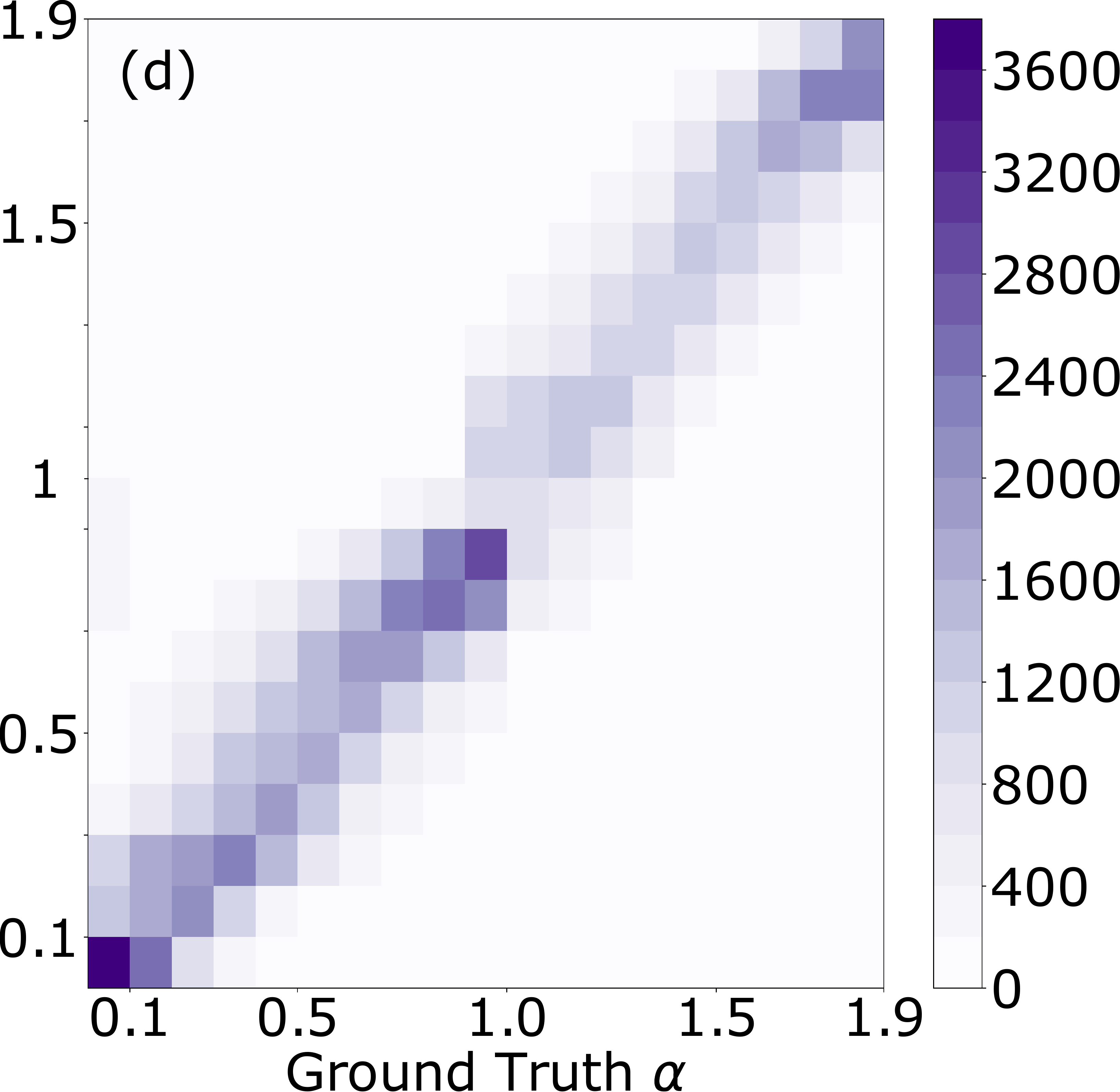}
  \caption{\label{fig:fig5} {\it Distribution of the predicted anomalous diffusion coefficient as a function of the ground truth, for all models.} In (a) and (b) we plot the distributions for $L=20$ and  $\rm{SNR}=2$ and   $\rm{SNR}=1$, respectively. In (c) and (d) we plot them for $L=500$ and again  $\rm{SNR}=2$ and   $\rm{SNR}=1$, respectively. } 
\end{figure} 

With regards to the f$_1$-score, we observe that there is an abrupt change at normal diffusion, where $\alpha=1$, which is more abrupt for long trajectories. We remark that not all models are used at all values of $\alpha$ (as we commented, ATTM and CTRW are restricted to $\alpha\le1$ and LW to $\alpha\ge1$). We note then that at $\alpha=1$ calculations are made with all five models.  Also, the results are slightly better for $\alpha>1$, which is also due to the existence of three models in this range, instead of the four models in the range $\alpha <1$. Also, notice that below $\alpha=1$ we consider the ATTM model, which as shown in Fig.~\ref{fig:fig3} has a lower accuracy. We also appreciate, that the longer the trajectory is, the higher the f1-score is. However, from $L=500$ up to $L=1000$ it seems that the gain is very small when increasing the trajectory length. Respect to the MAE, the results are more or less stable around   $\alpha=1$ and get worse for shorter trajectories. Nevertheless, it shows pronouncedly  that for short trajectories and closer to $\alpha=0$ and $\alpha=2$ the error increases. This also occurred in the figures for f$_1$-score, but only for $\alpha <1$.  For longer trajectories this effect is reduced and even inverted close to  $\alpha=2$.

We plot in  Fig.~\ref{fig:fig5}   the distribution of $\alpha$ predicted as a function of the ground truth $\alpha$, for the pool of models and two different lengths and noise levels. Since the dataset is generated with an equal number of trajectories of each model, and there are 4 models in the subdiffusive regime and 3 in the superdiffusive one, there are more trajectories in the left part of the pictures. The spread around the correct value in the diagonal is similar at both sides of $\alpha=1$. Also, the distribution is wider for shorter trajectories than for longer trajectories (top panels versus bottom panels). Noise seems not to have a large impact in the results shown here. Finally, the decrease in the performance around $\alpha=1$ agrees with the results in Figure \ref{fig:fig4}.

\begin{figure}
 \includegraphics[width=0.49\columnwidth]{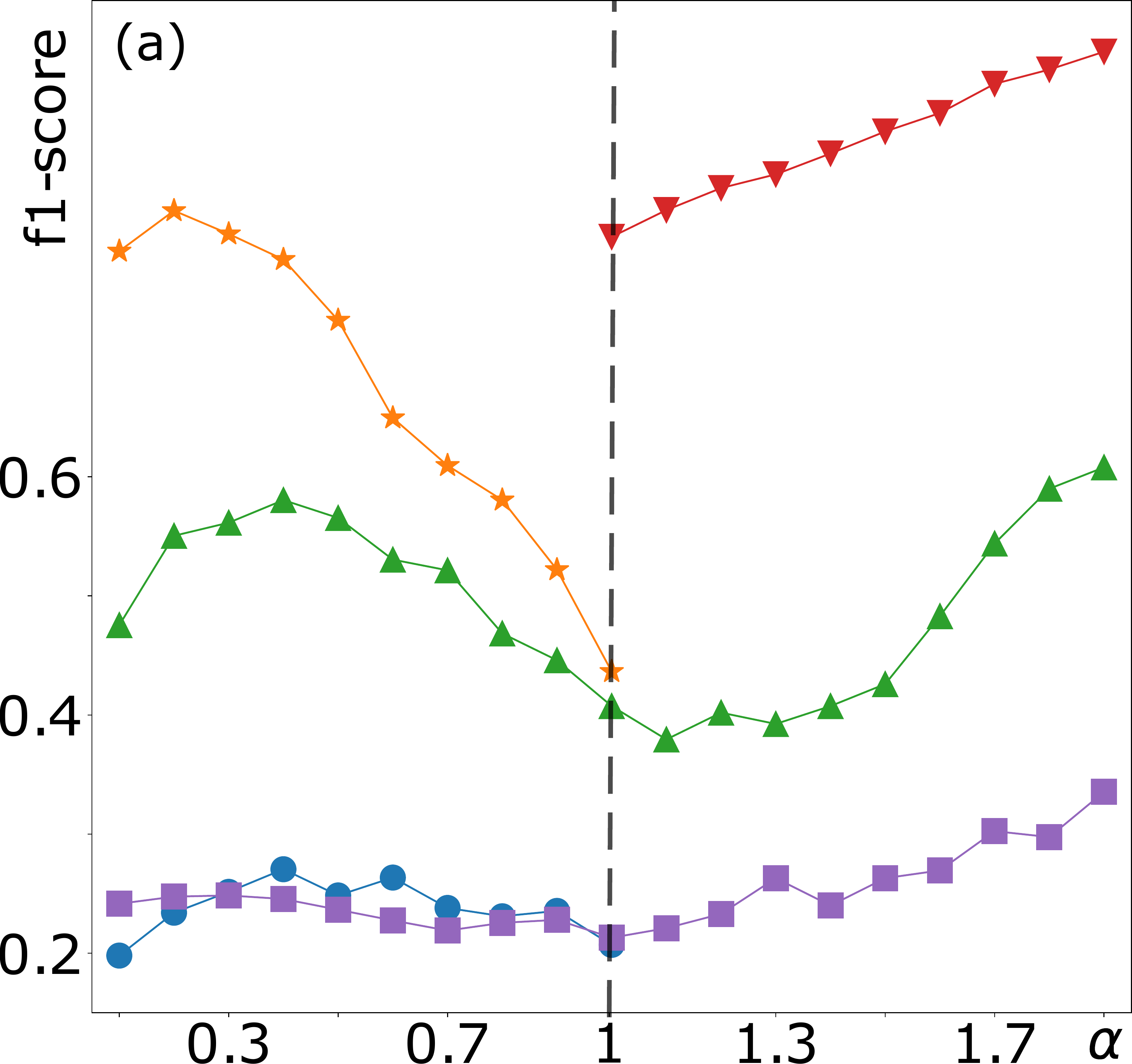} 
 \includegraphics[width=0.49\columnwidth]{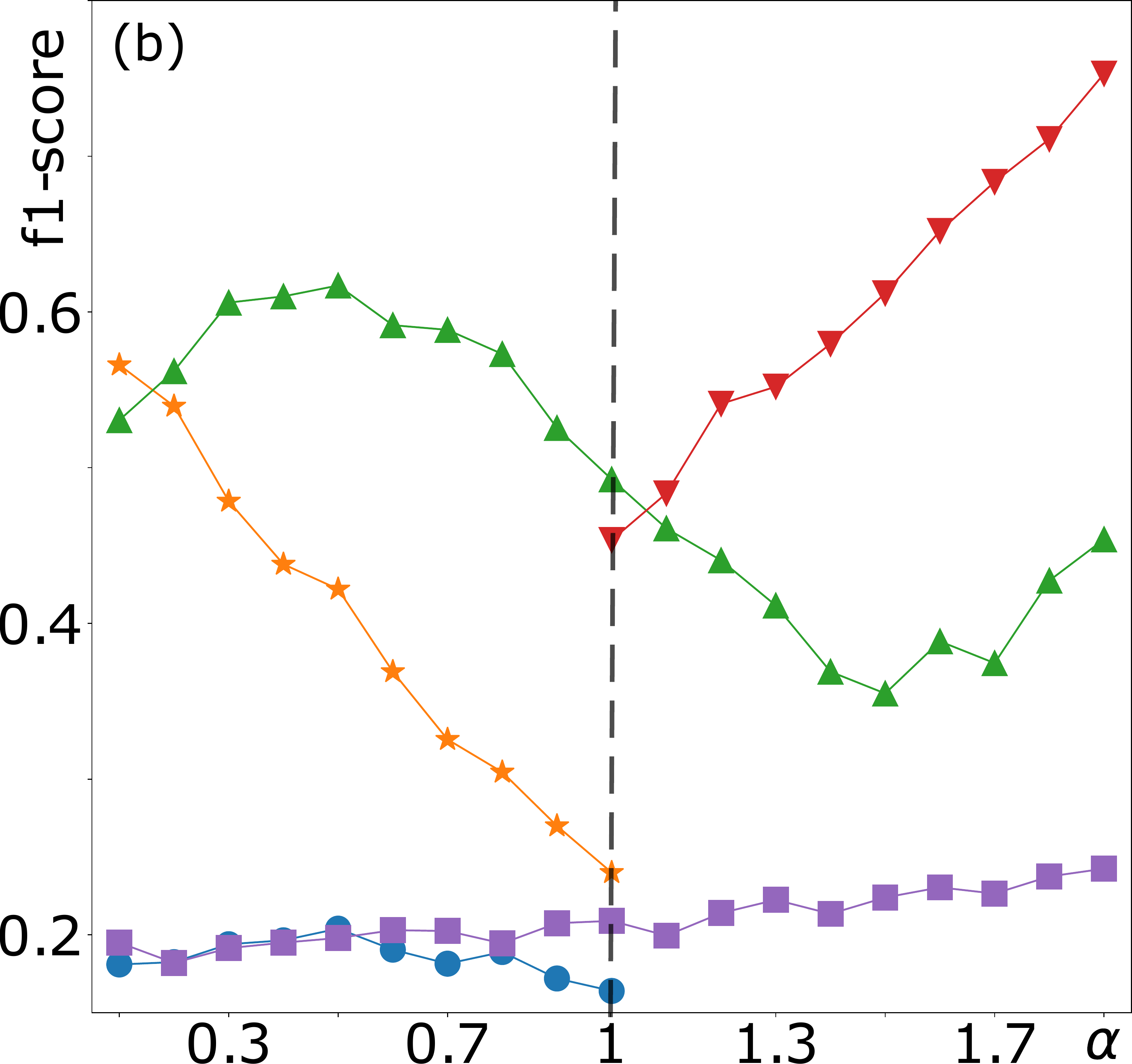}
\vspace{-0.2cm}
 \includegraphics[width=0.49\columnwidth]{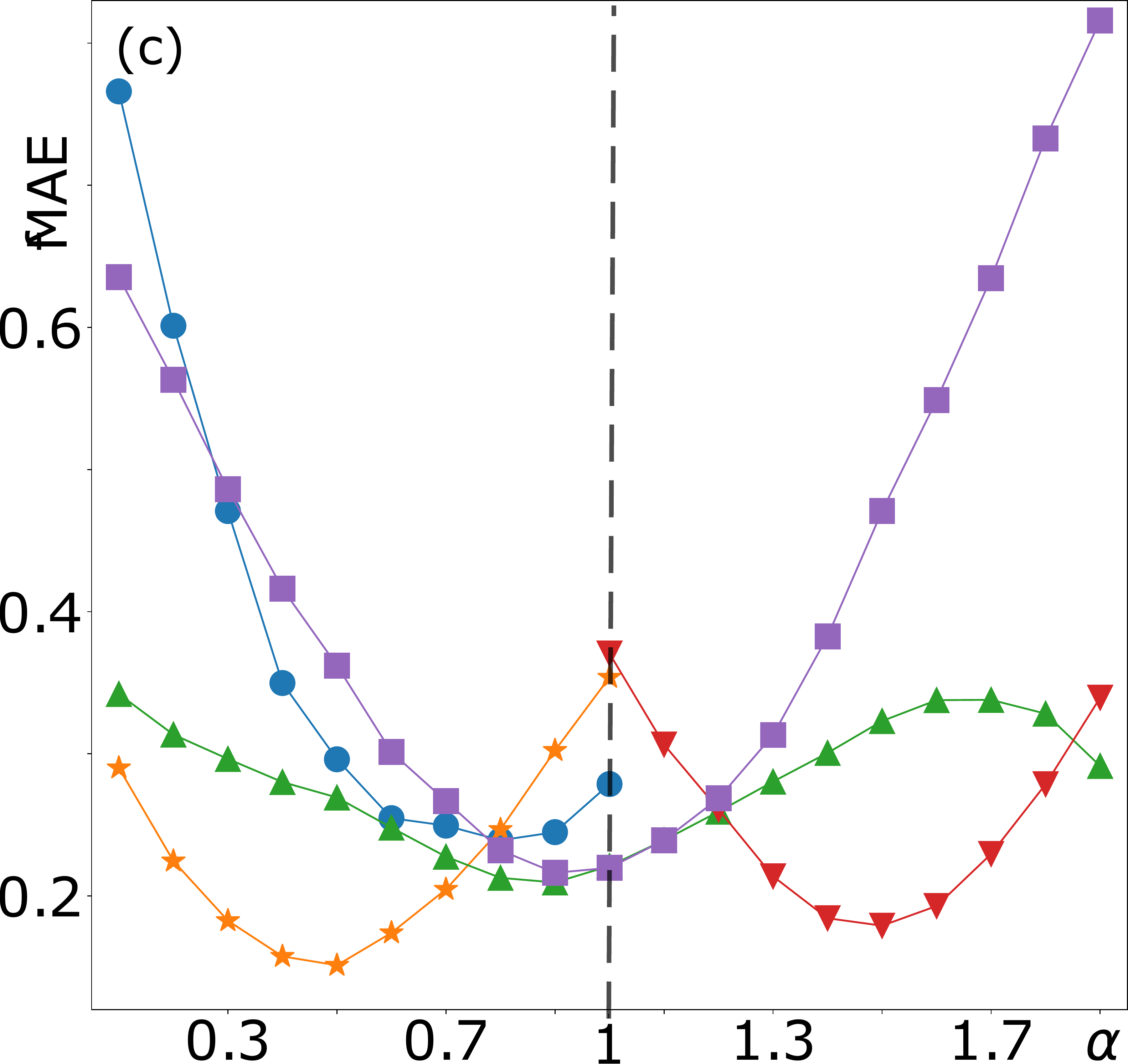} 
 \includegraphics[width=0.49\columnwidth]{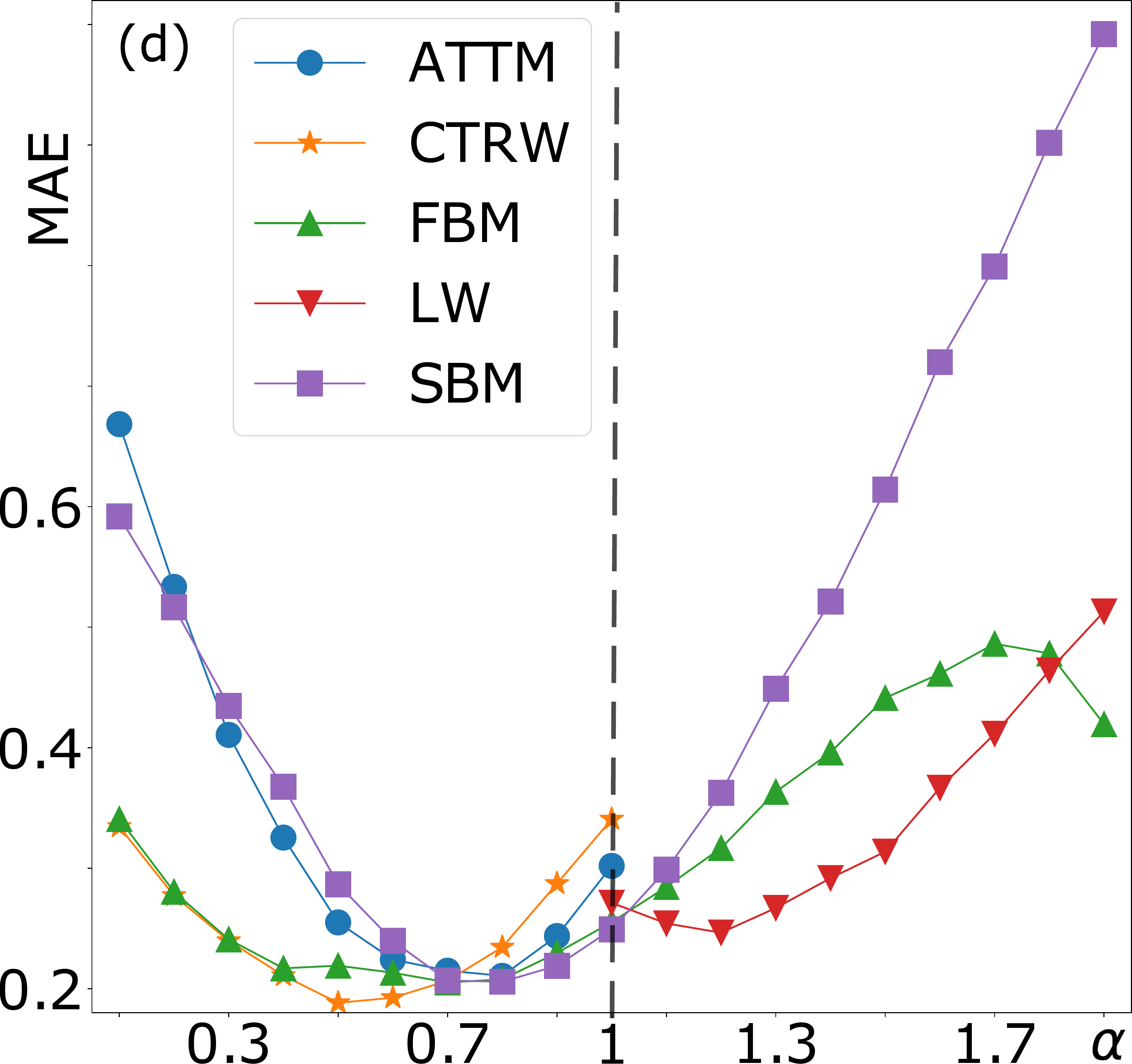}
 \centering
 \caption{\label{fig:fig6} {\it Inference of the anomalous exponent $\alpha$ for different values of the ground truth and for different theoretical models.}  For $L=20$, F$_1$-score and MAE for different values of ground truth anomalous exponent for $\rm{SNR}=2$ in (a) and (c); and the same for  $\rm{SNR}=1$ (b) and (d).}
\end{figure}

\begin{figure}
 \includegraphics[width=0.49\columnwidth]{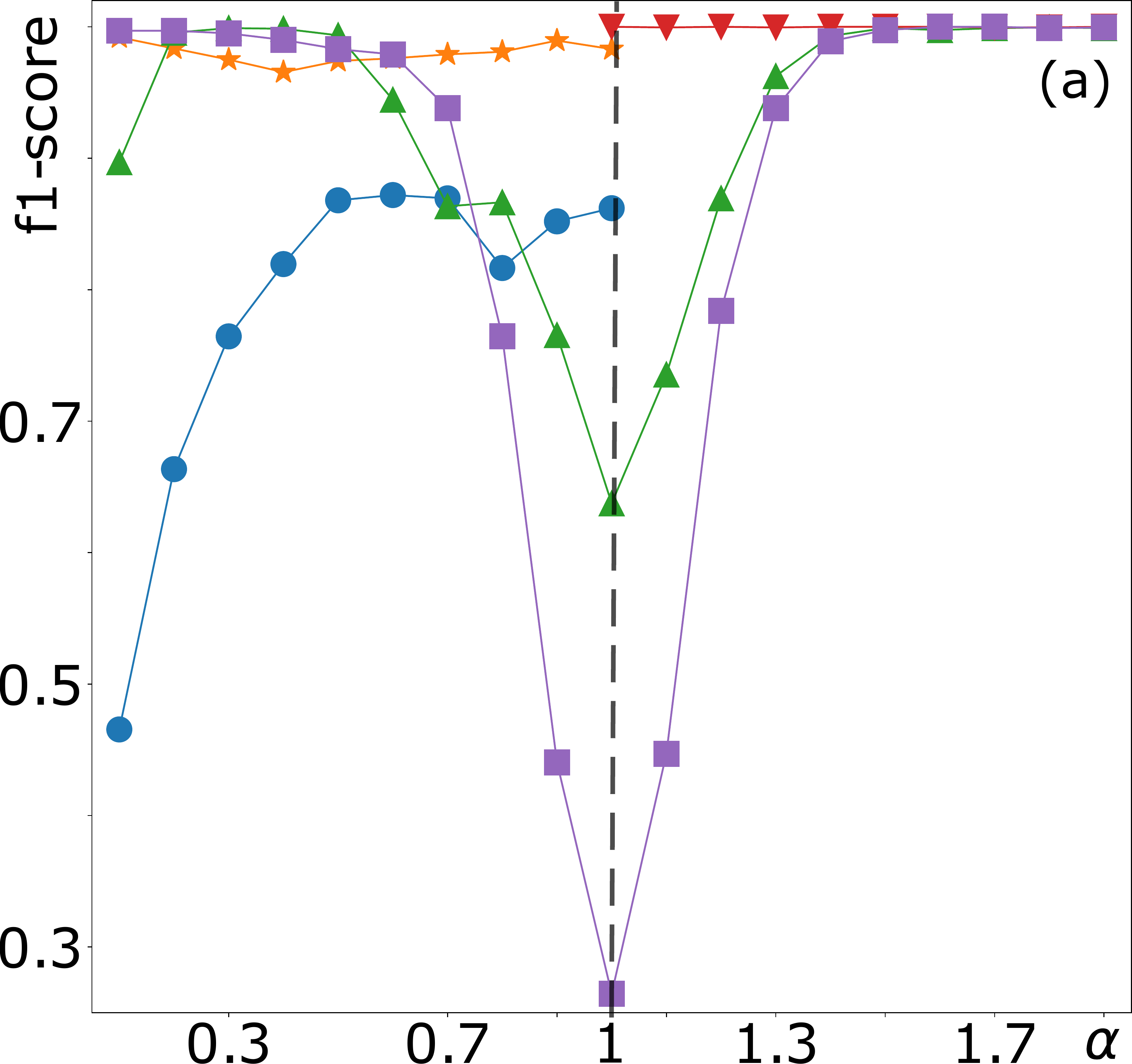} 
 \includegraphics[width=0.49\columnwidth]{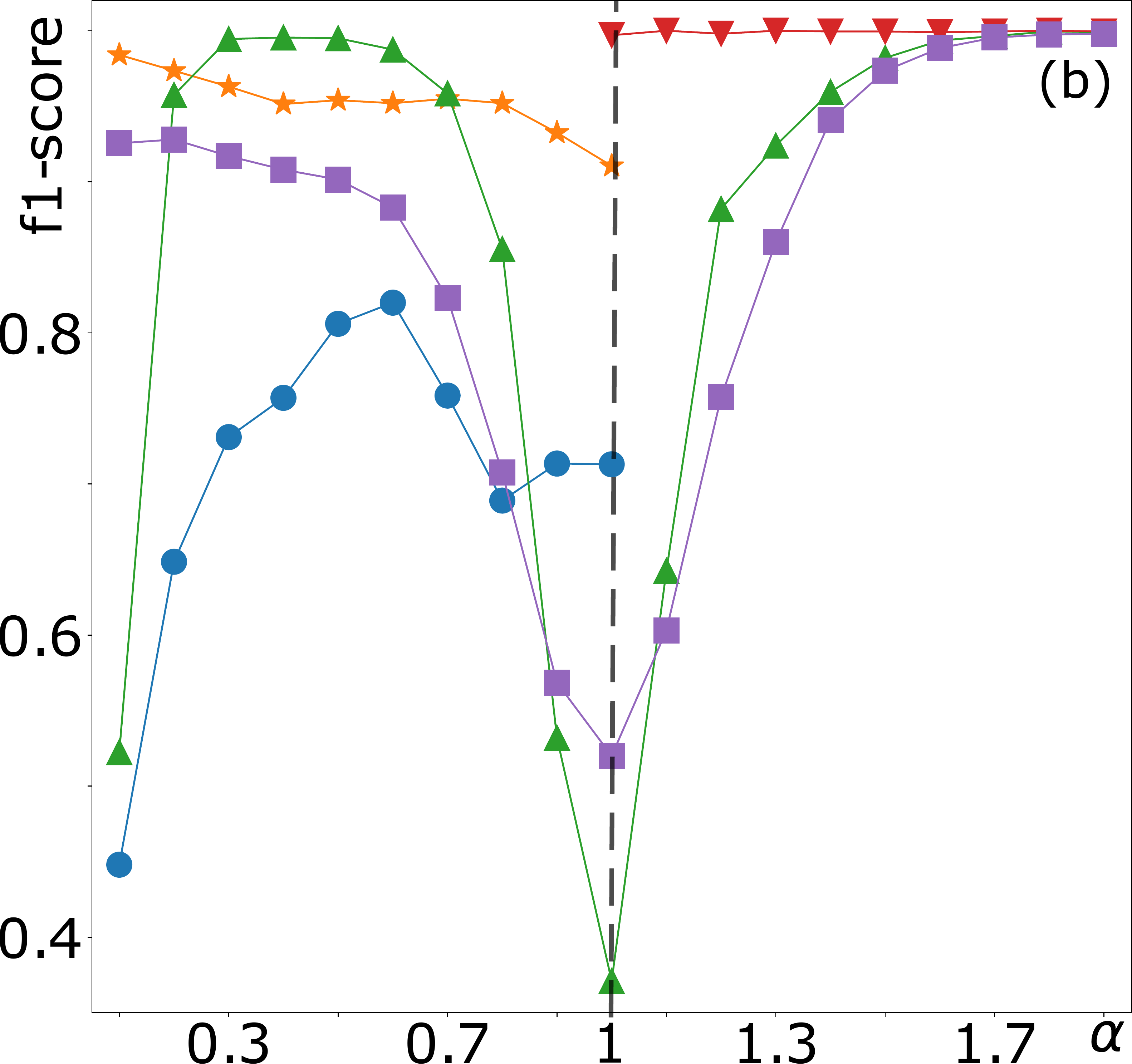}
 \includegraphics[width=0.49\columnwidth]{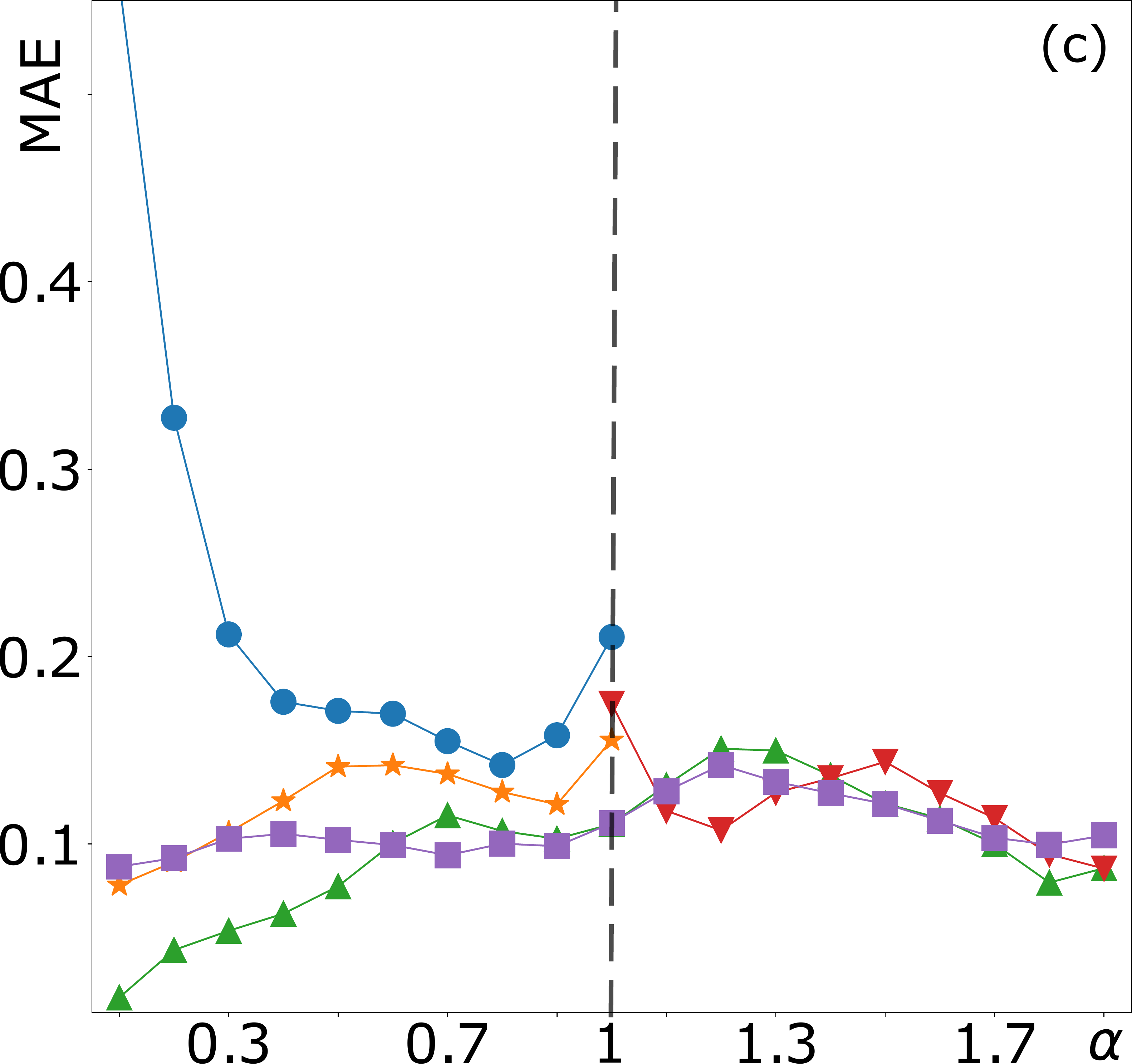} 
 \includegraphics[width=0.49\columnwidth]{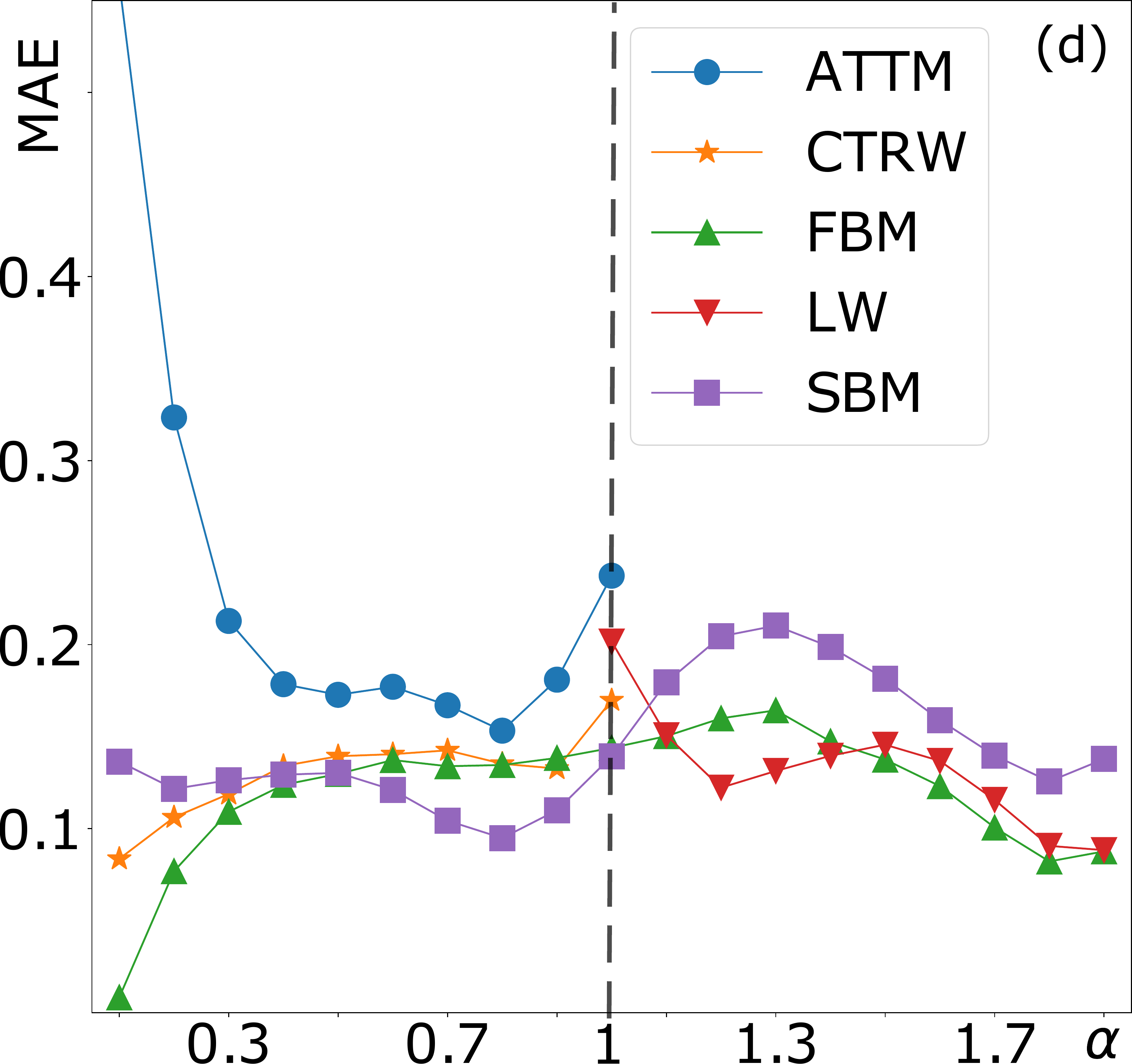}
 \centering
 \caption{\label{fig:fig7} {\it Inference of the anomalous exponent $\alpha$ for different values of the ground truth and for different theoretical models.}  (a)  F$_1$-score and (c)  Mean absolute error   for different values of ground truth anomalous exponent for  $L=500$ and $\rm{SNR}=2$; (b) and (d) same for  $\rm{SNR}=1$.}
\end{figure}

To get further insight, we fix the length and SNR and we plot, for each model, the MAE and   f$_1$-score for different values of the $\alpha$  in the interval   $[\alpha_{\rm{GT}}-\Delta \alpha/2,\alpha_{\rm{GT}}+\Delta \alpha/2]$, for $L=20$  and two different levels of noise (see Figs.~\ref{fig:fig6}  and~\ref{fig:fig7}). Notice that the range of $\alpha$ covered by each model is different.

In Fig.~\ref{fig:fig6}, we observe that the anomalous exponent inference of ATTM and SBM is poorer that in the rest of models. Similarly, the MAE gets worse as long as we approach to the limits of the exponent range:  $\alpha=0$ and $1$ for ATTM and $\alpha=0$ and $2$ for SBM.
On the contrary, the CTRW and LW, that they not cover the whole exponent range, respond better when approaching to the limits away different from $\alpha=1$. Lastly, the anomalous exponent is easier to identify for the FBM in the subdiffussive regime. 

In Fig.~\ref{fig:fig7} we plot the same for $L=500$.  Here, the better predictions are obtained for the CTRW and LW models. The worst results are exhibited for the FBM and SBM around the normal diffusion regime. This is probably due to the fact that around $\alpha=1$ we can find trajectories of all models, which confuses the predictions of the models that really exhibit trajectories with this diffusion exponent.
We also see that the performance of the FBM decreases a lot around $\alpha=0$ with high noise. Lastly, an increase of the length neither helps to improve the performance of the ATTM around $\alpha=0$.

\begin{figure}
 \includegraphics[width=0.49\columnwidth]{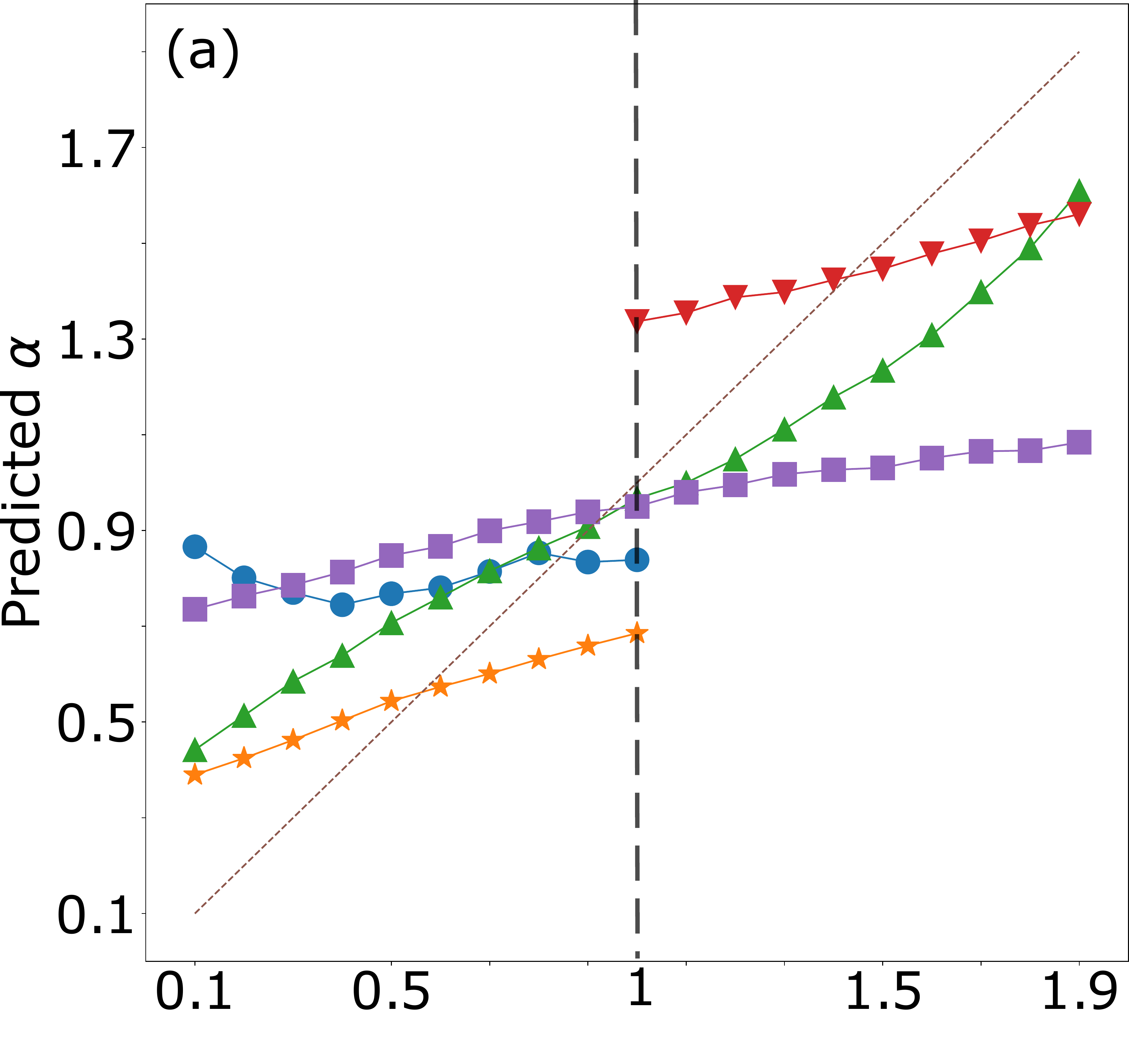}
 %\centering
 \includegraphics[width=0.49\columnwidth]{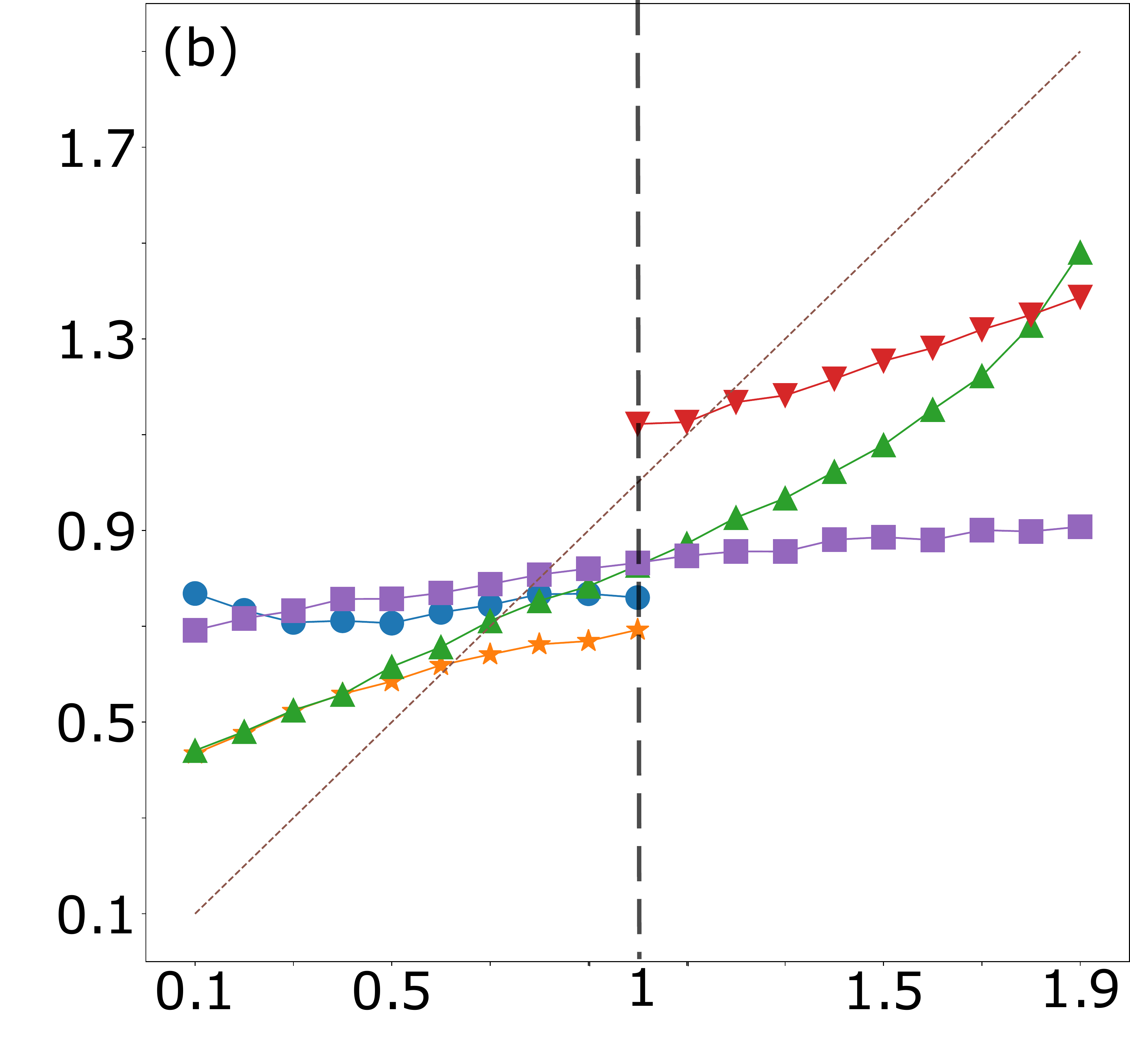}\\
 \includegraphics[width=0.49\columnwidth]{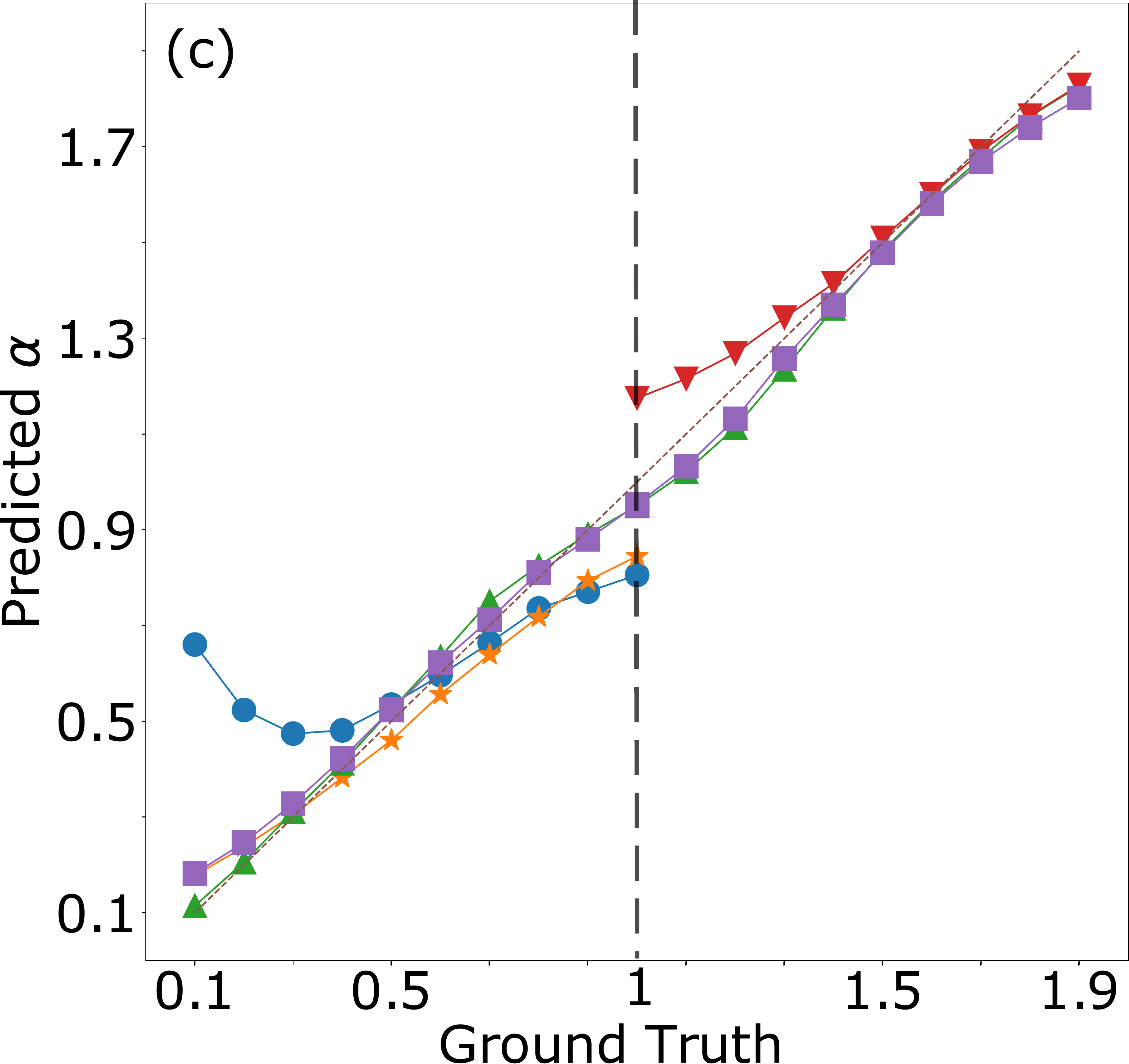}
 %\centering
 \includegraphics[width=0.49\columnwidth]{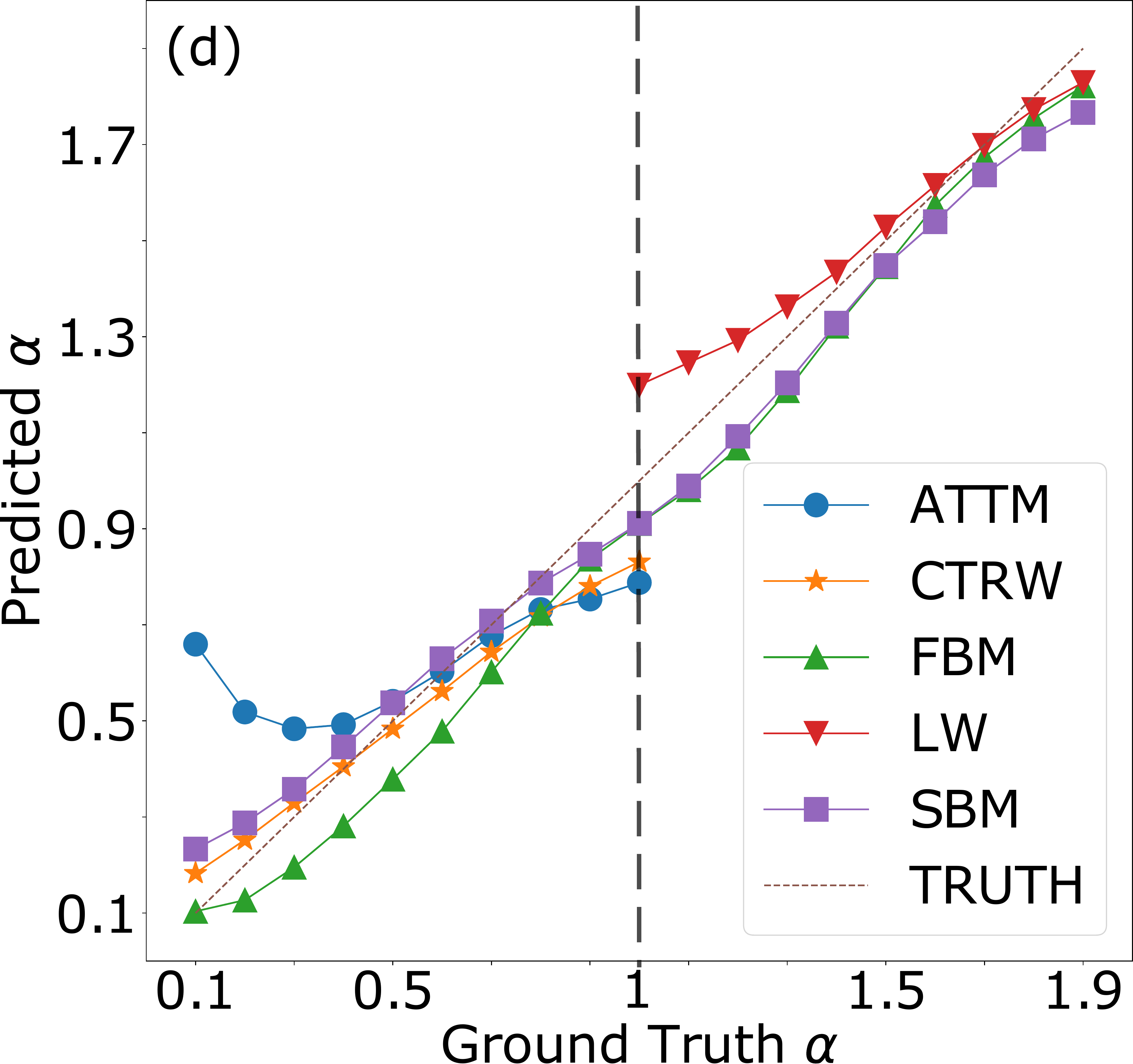}
 \caption{\label{fig:fig8} {\it Average of the predicted amomalous diffusion coefficient as a function of the ground truth, for each models.} In (a) and (b) we plot the average  for $L=20$ and  $\rm{SNR}=2$ and   $\rm{SNR}=1$, respectively. In (c) and (d) we plot them for $L=500$ and again  $\rm{SNR}=2$ and   $\rm{SNR}=1$, respectively.  }
\end{figure}

We finally plot in  Fig.~\ref{fig:fig8}  the average of the $\alpha$ predicted as a function of the ground truth $\alpha$ for each model and two different lengths and noise levels, which disaggregates the results shown in Figure \ref{fig:fig7}. Interestingly, in short trajectories, the predicted values of $\alpha$ in the subdiffusive regime tend to be higher than the ground truth. Conversely, in the superdiffusive regime, they tend to be slightly smaller ground truth. Nevertheless, we only find a clear bias for long trajectories in the ATTM model, which tends to predict smaller values than the real ones close to $\alpha=0$, and in the $LW$ around $\alpha=1$.

\subsection{Diffusion model classification}

\begin{figure}
 \includegraphics[width=0.49\columnwidth]{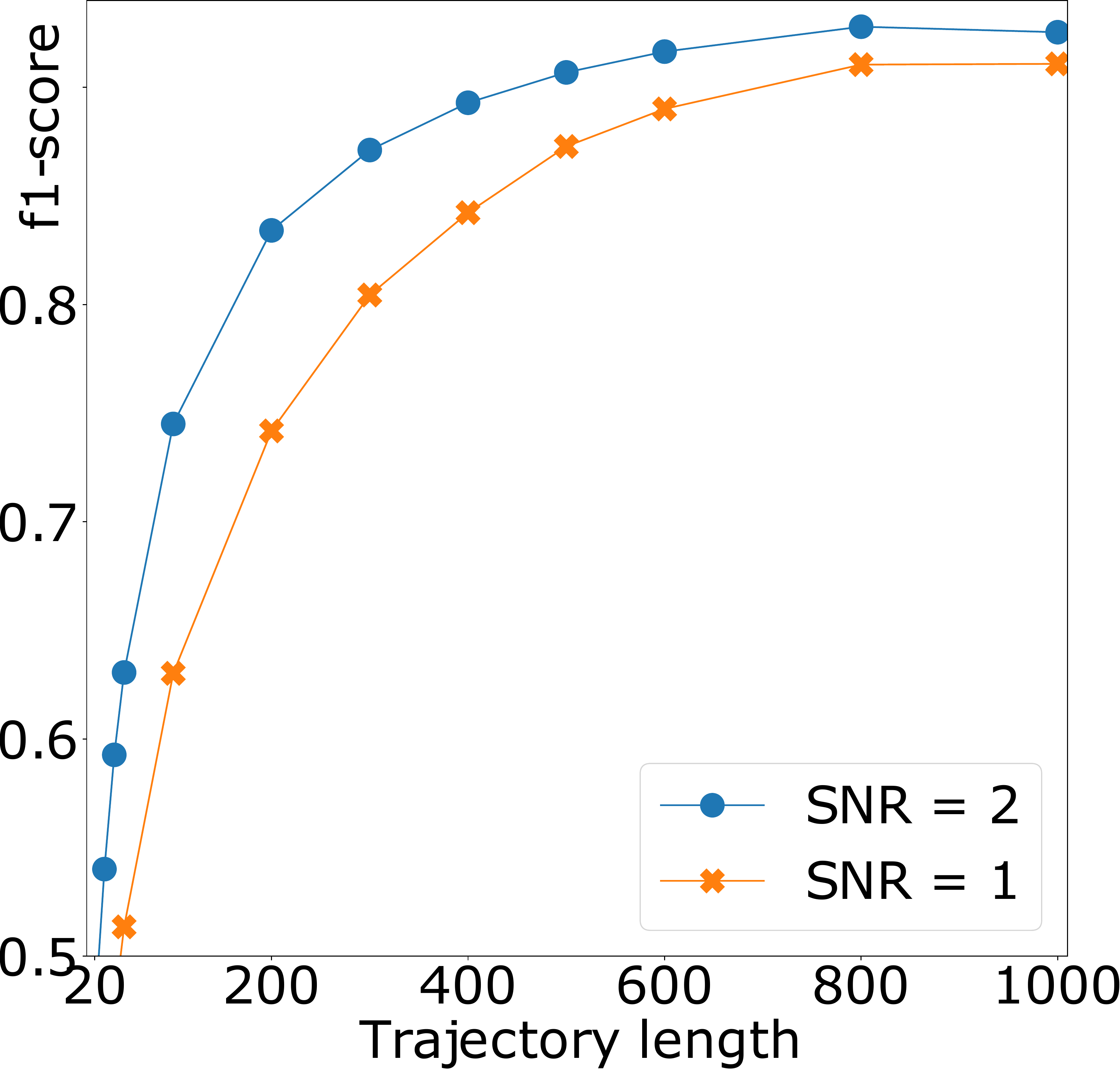}
 \centering
 \caption{\label{fig:fig9} {\it Diffusion model classification as a function of length} f$_1$ score as a function of length for to levels of noise in one  dimension.}
\end{figure}

The second task is to predict the model which explains better the  trajectory at hand. In Fig. \ref{fig:fig9} we plot the f$_1$ score as a function of trajectory length, showing the expected behavior (better results for longer trajectories and less noise). Similarly to the previous task there is a stabilization of the improvement of  f$_1$ score  around $L=400$. 

\begin{figure}
 \centering
 \includegraphics[width=0.49\columnwidth]{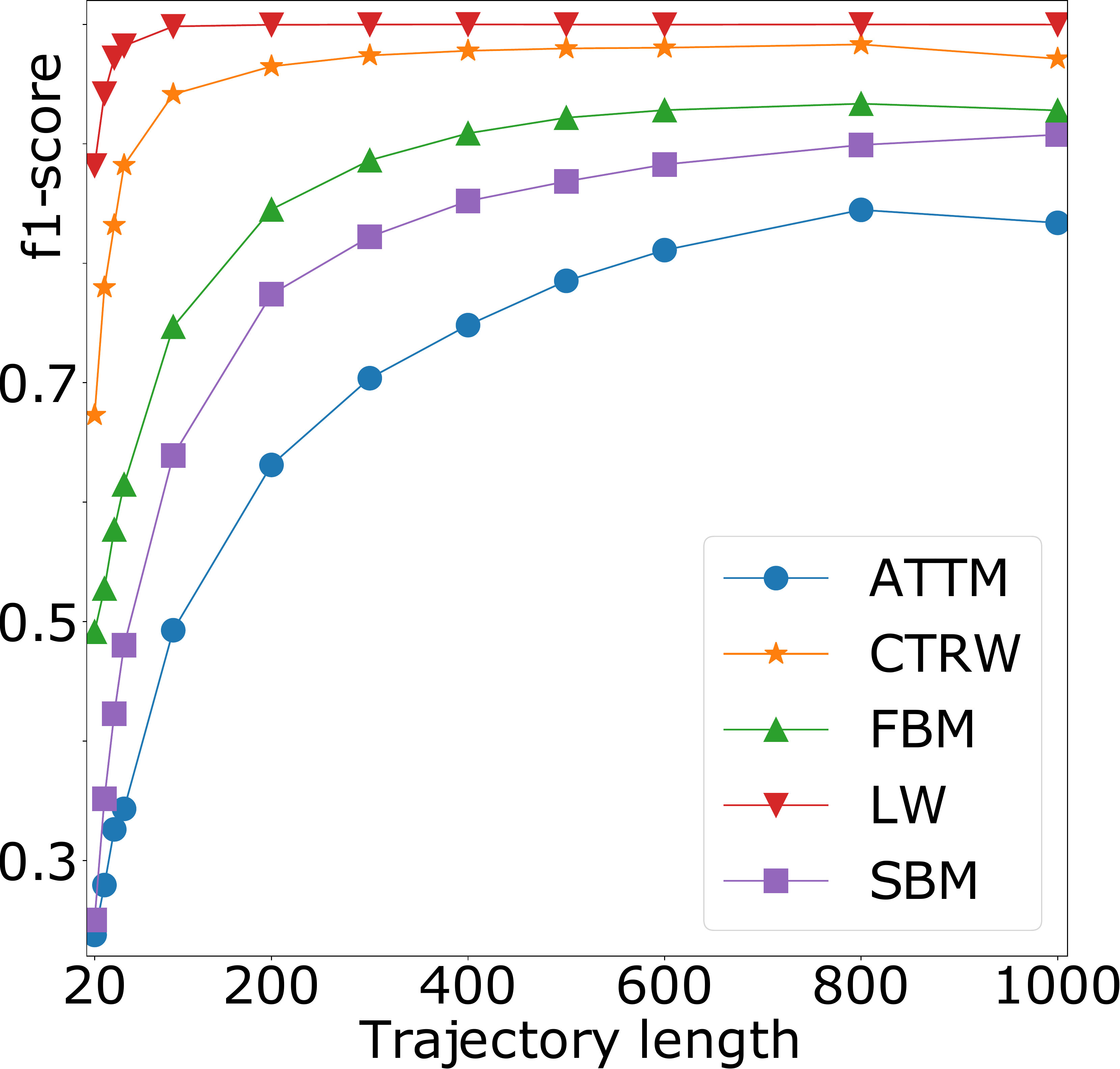}
 \vspace{0.5cm}
 \includegraphics[width=0.49\columnwidth]{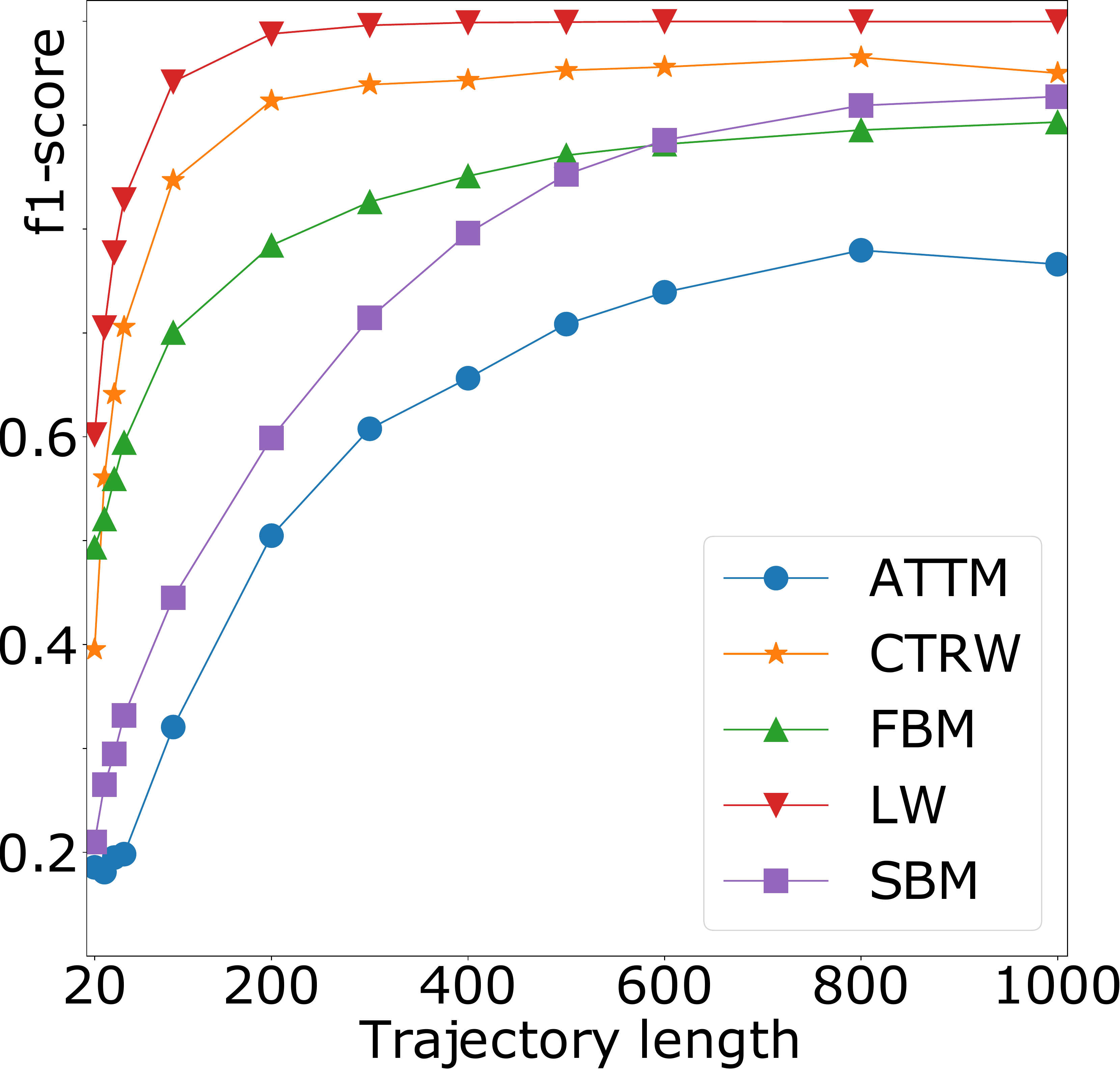}
 %\centering
 \caption{\label{fig:fig10} {\it Diffusion model classification as a function of length, for different models} F$_1$ score as a function of length for (a) $\rm{SNR}=2$ and (b)  $\rm{SNR}=1$  for one dimension.    }
\end{figure}

We also plot  the f$_1$ score as a function of trajectory length for the different models and two noise levels in Fig. \ref{fig:fig10}. Here, one can observe that the f$_1$ score is always larger for LWs and stabilizes in a large value for both levels of noise even at shorter lengths. Again, ATTM is the one that behaves the worst for all lengths. Finally, SBM also behaves worst than all other models except ATTM but, reminiscently of previous task, it improves faster than the rest of models for the larger level of noise.  

\begin{figure*}
\centering
 \includegraphics[width=0.49\columnwidth]{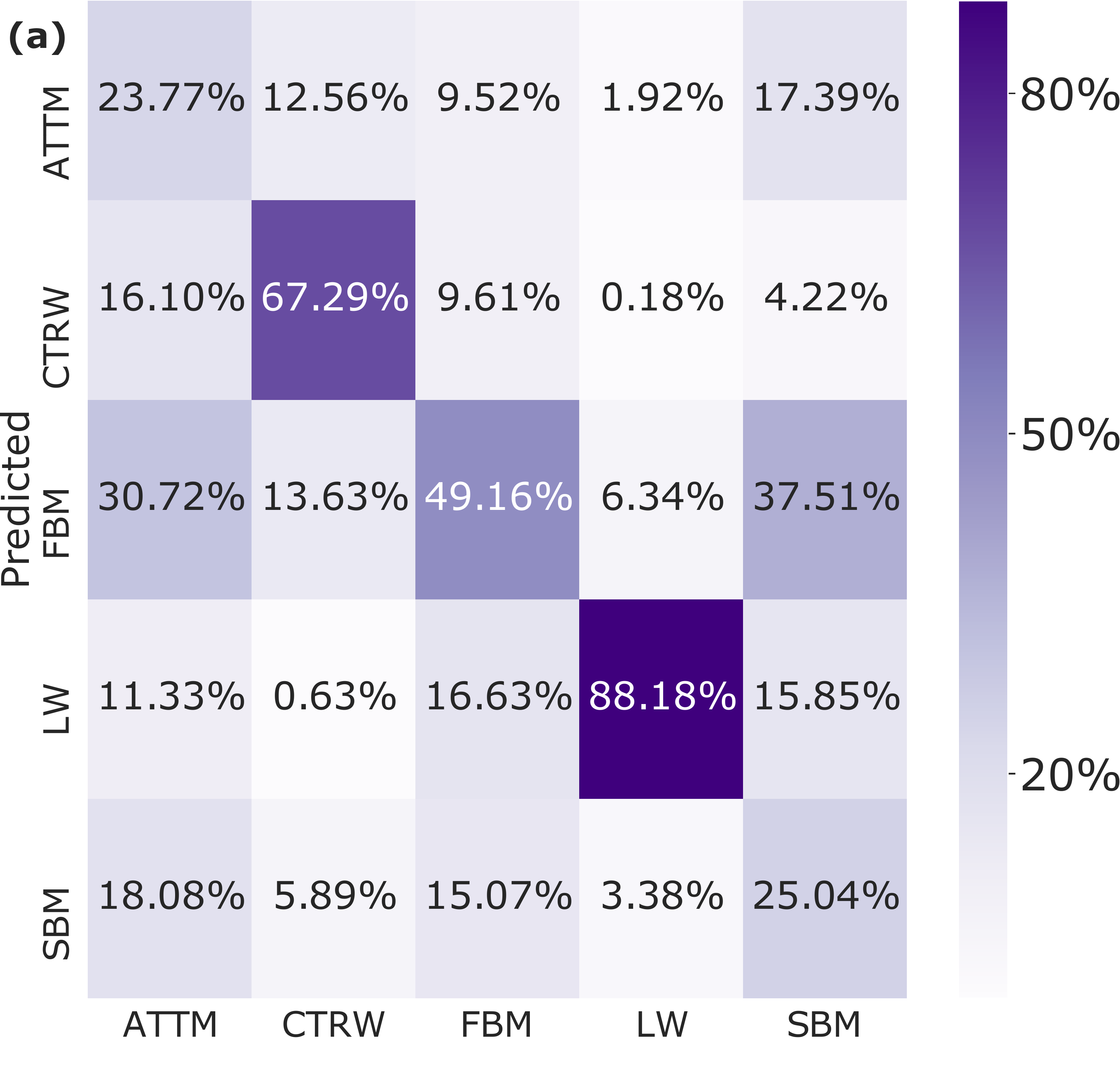}
 \includegraphics[width=0.49\columnwidth]{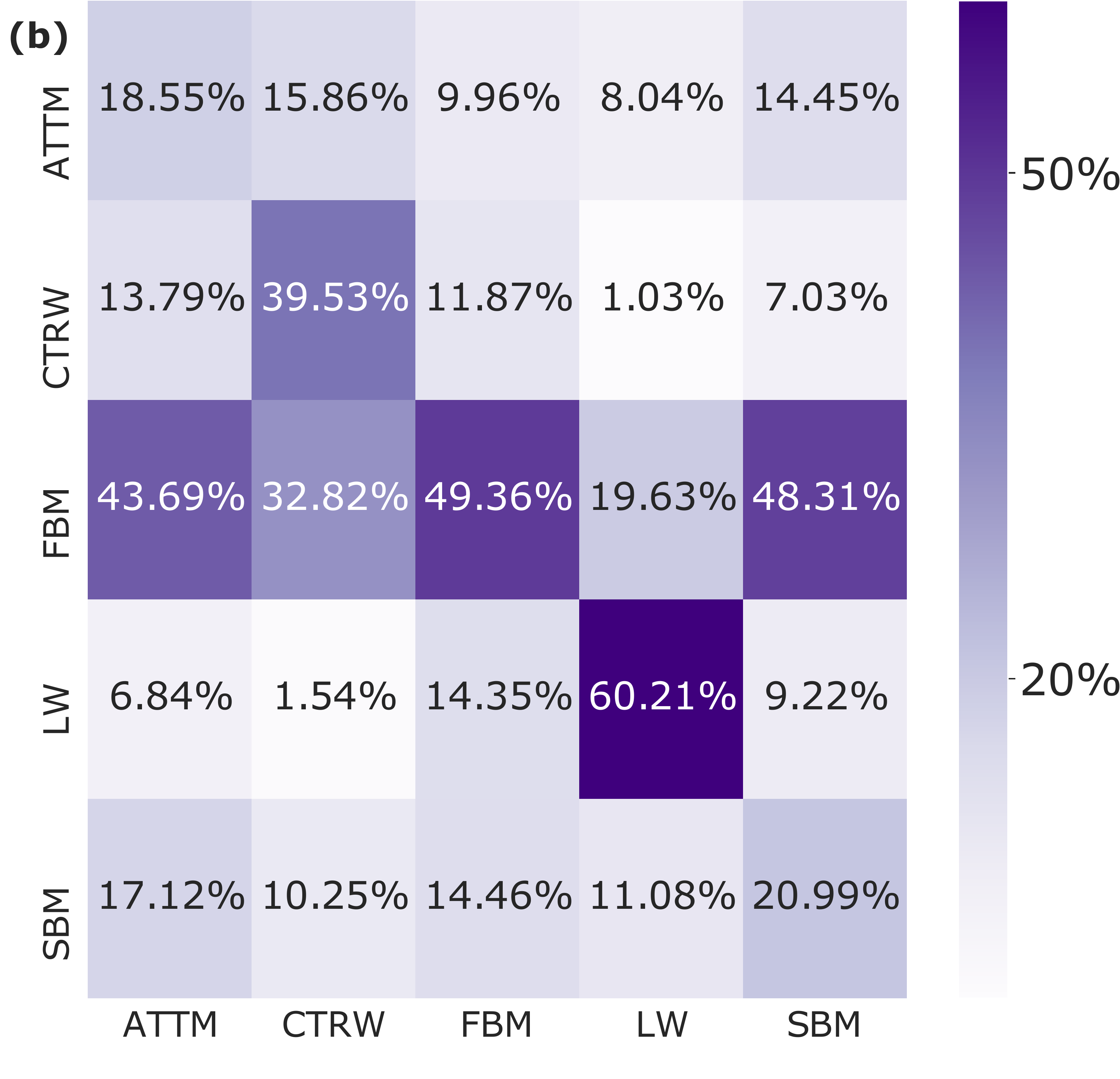}\\
 \includegraphics[width=0.49\columnwidth]{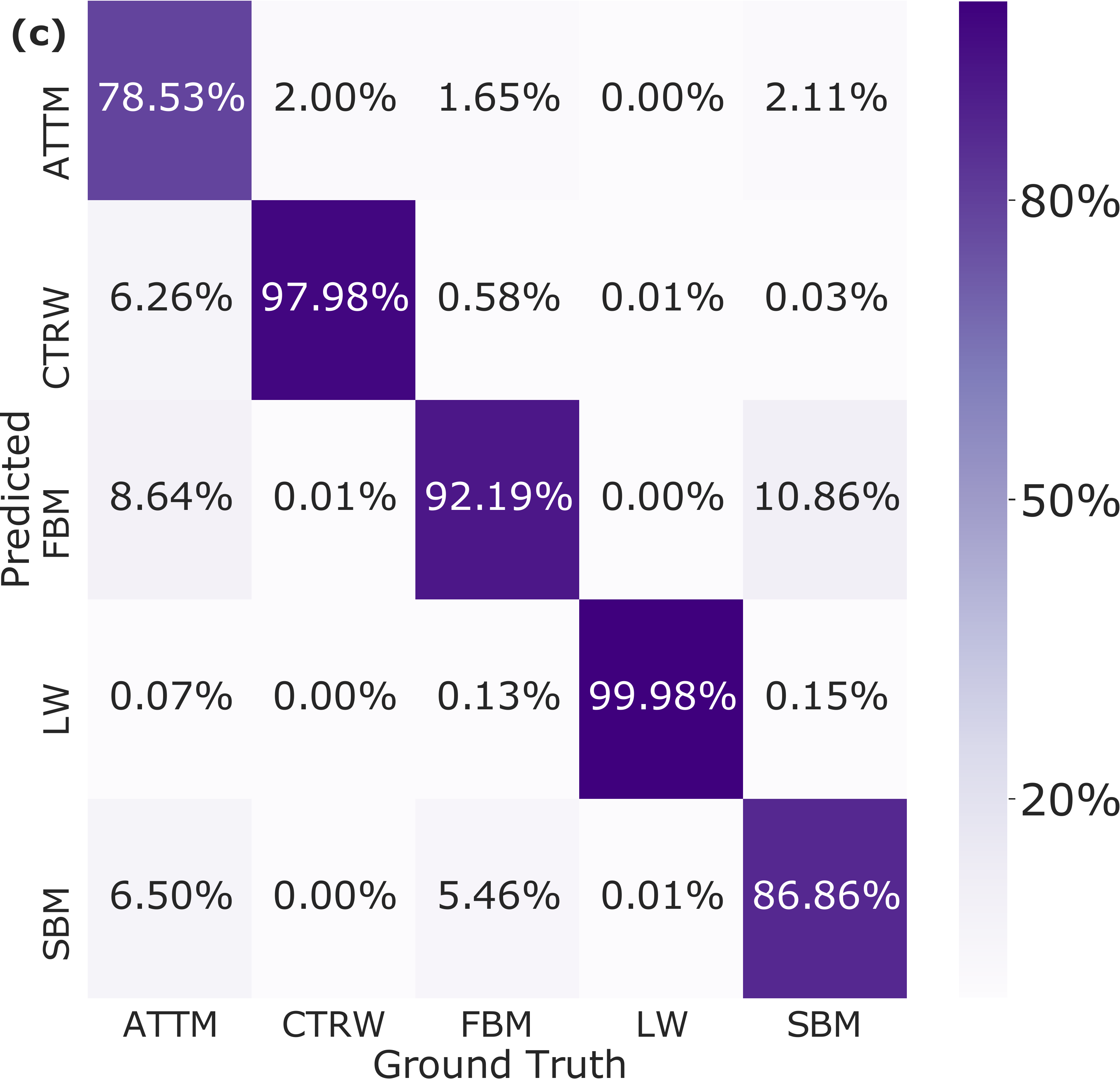}
 \includegraphics[width=0.49\columnwidth]{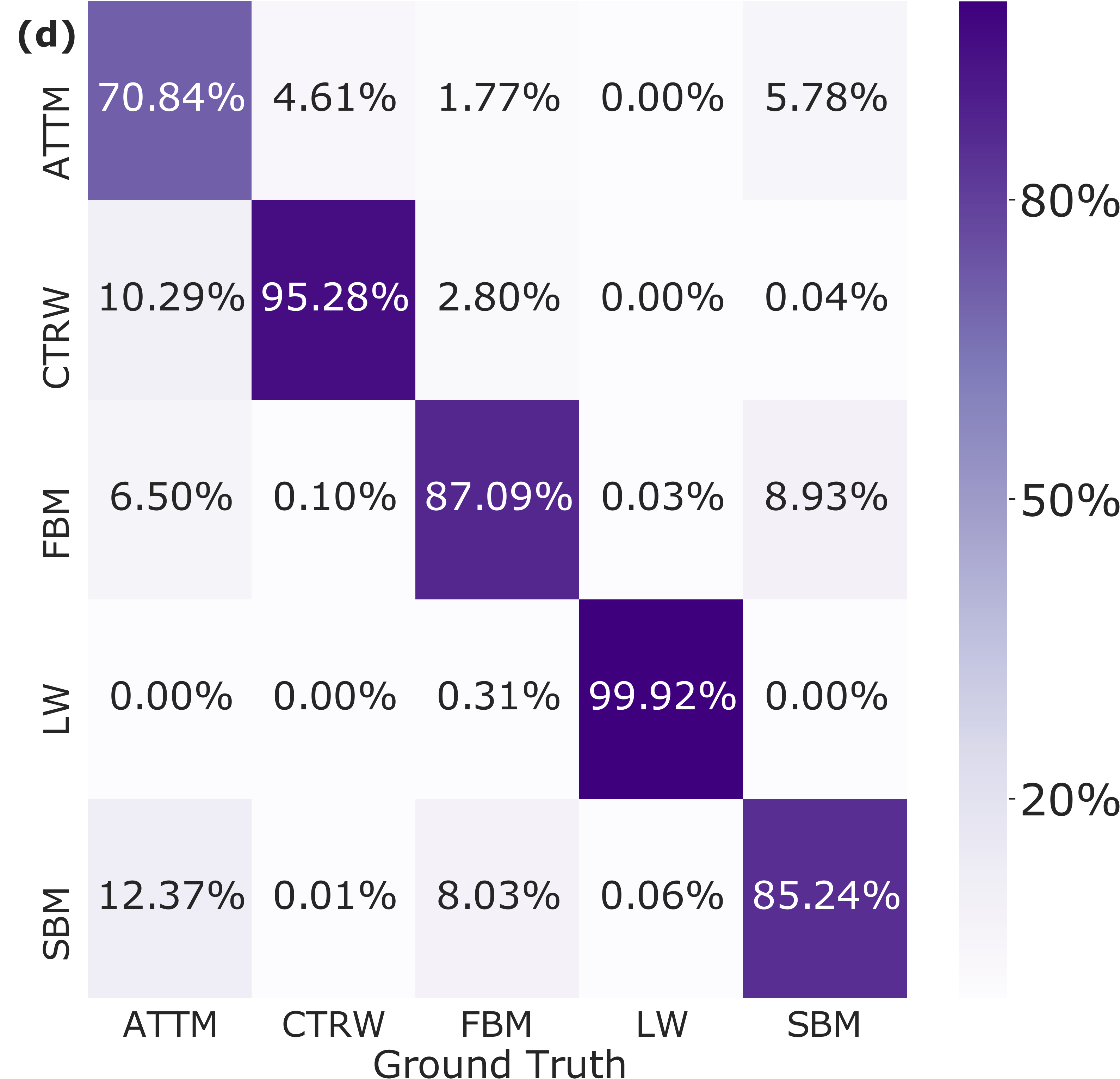}
 %\includegraphics[width=0.5\columnwidth]{Fig9c}
 %\centering
 %\includegraphics[width=0.5\columnwidth]{Fig9d}\\
 %
 %\centering
 \caption{\label{fig:fig11} {\it Confusion matrices,} calculated for  $L=20$ with SNR$=2$  (a) and  SNR$=1$ (b),  and   for  $L=500$ with SNR$=2$  (c) and  SNR$=1$ (d).}
\end{figure*}

We plot some exemplary confusion matrices in Fig. \ref{fig:fig11}, for different lengths and noise levels. As shown before, the identification of the LW is very clear, even in short trajectories. In this case, the short trajectories of the rest of the models are often confused with the FBM. Again, the CTRW and FBM are more accurately classified than the ATTM and SBM models. Finally, for long trajectories, we also observe that the ATTM is the worst classified method. We also note that the performance of the SBM was equally bad as the ATTM in short trajectories but increases quite a lot when the trajectories length increases.

Finally, we note that the results showed in Fig.~\ref{fig:fig2} gave some hints on the model classifier. As commented, for moderate noise, $SNR=2$, ATTM and  SBM follow the same behavior (abrupt change below $L=300$ and stabilization for $L>300$). As we see now, the SBM is harder to identify with smaller noise than ATTM in short trajectories. However, the model classifies a little better ATTM.   We can see that for short trajectories, with a larger SNR, the confusion between ATTM and SBM decreases. 

\section{Conclusions}\label{sec:conc}

We presented the computational tool we used to participate in the AnDi challenge, which took place in 2020 (\url{http://www.andi-challenge.org}) \cite{2020MunozArxiv,2021munoz-gilArx}. In the challenge the tools presented here ranked among the top 4 in all tasks, being the best in the regression task in one dimension. Besides, in the classification task, it ranked fourth in one dimension. With a similar model used with vectors containing all the trajectory coordinates,  the model ranked third in two dimensions and second in three dimensions. The tool is a combination of  convolutional and recurrent neural networks based on bidirectional LSTM blocks.ional layers, and we 

For task 1, i.e. inferring the anomalous exponent for a single trajectory, we obtain good MAEs below 0.2 for trajectories over length $L=200$,  even for the largest level of noise considered here. Also, for shorter trajectories we obtain reasonably good MAEs, finding a limit around $L=50$. No further information is supposed to be at disposal when analyzing a single  trajectory, if one wants to infer the associated anomalous coefficient. Anyhow, it is illustrative to study whether the MAE is different if we test the model only with trajectories generated with one model. We found that the worst performing for short trajectories are ATTM and SBM. Also, it is informative to see if the behavior error is larger if the analyzed trajectory has an $\alpha$ close to say one or zero. We found that error is larger close to normal diffusion, as expected, and in some cases close to the limit of very trapped trajectories ($\alpha=0$) and close to ballistic motion ($\alpha=2$).  Also, we showed that the dispersion on the values of $\alpha$ predicted is larger for shorter trajectories, as expected. 

For task 2, i.e. classification of trajectories, the code should be able to assign a theoretical model to a given trajectory with large accuracy. We found that again we are able to obtain f$_1$-scores above 0.8 for long enough trajectories, with  the accuracy dropping down for shorter trajectories  and finding again a limit around $L=50$. Again, no further information is supposed to be associated to the trajectory. But for academic information, we studied how f$_1$-score changes if we consider only trajectories of one of the models. We found  again that the worst behaving models are ATTM and SBM. Finally, the confusion matrices show LW is easily identified and  not confused with other models. This is to be expected as this model has peculiarities very different to other models, i.e. the correlation between step length and time waited. Also,  all models are often confused with FBM. Finally,  CTRW and FBM are more accurately classified than the ATTM and SBM models, and for  long trajectories,  the ATTM is the worst classified method.

In summary, the tool presented here offers good accuracies in both tasks. Dimensions two and three have not been discussed here, but the results are shown in \cite{2021munoz-gilArx}. The model shows, in higher dimensions, similar performances as the best performing methods in the challenge. As an outlook, we aim at using this tool in the task 3 of AnDi-Challenge, where one has a trajectory which changes behavior (anomalous coefficient or diffusion model) in an intermediate point, and the goal is to find accurately this point. First trials with our tool showed good results, and we will explore this in the future. 

\section*{Acknowledgements}
J.A.C. acknowledges support from ALBATROSS project (National Plan for Scientific and Technical Research and Innovation 2017-2020, No. PID2019-104978RB-I00). 
M.A.G.M. acknowledges funding from the Spanish Ministry of Education and Vocational Training (MEFP) through the Beatriz Galindo program 2018 (BEAGAL18/00203) and Spanish Ministry MINECO (FIDEUA PID2019-106901GBI00/ 10.13039/501100011033).

%\ack We thank our families. 

%\appendix 

 %\section{Confusion matrices}
 %\label{sec:app} 

 %\MAGM{The distribution of this and that is calculated like this, and we plot it in \ref{fig:fig8}. The confusion matrices are calculated like that and we plot them in \ref{fig:fig9}} \MAGM{anything else to discuss here?}

%\bibliographystyle{unsrt}
%\bibliography{biblio.bib}
%\printbibliography

\end{document}